\documentclass{article}

\usepackage{arxiv}

\usepackage[utf8]{inputenc} % allow utf-8 input
\usepackage[T1]{fontenc}    % use 8-bit T1 fonts
\usepackage{hyperref}       % hyperlinks
\usepackage{url}            % simple URL typesetting
\usepackage{booktabs}       % professional-quality tables
\usepackage{amsfonts}       % blackboard math symbols
\usepackage{nicefrac}       % compact symbols for 1/2, etc.
\usepackage{microtype}      % microtypography
\usepackage{lipsum}		% Can be removed after putting your text content
\usepackage{graphicx}

\usepackage{amsthm,amssymb}
\usepackage{amsmath}
\usepackage{algorithm2e}
\usepackage{graphicx}
\usepackage{todonotes}
\usepackage{subcaption}
\usepackage{multirow}
\usepackage{tikz}

\title{Supervised Dimensionality Reduction and Visualization using Centroid-encoder}

\author{ \href{https://www.cs.colostate.edu/~tomojit/}{Tomojit Ghosh} \\
	Department of Computer Science\\
	Colorado State University\\
       Fort Collins, CO 80523 ,USA \\
	\texttt{Tomojit.Ghosh@colostate.edu} \\
	\And
	\href{https://www.math.colostate.edu/~kirby/}{Michael Kirby} \\
	Department of Department of Mathematics\\
       Colorado State University\\
       Fort Collins, CO 80523, USA \\
	\texttt{Kirby@math.colostate.edu} \\
}

\begin{document}
\maketitle

\begin{abstract}
	Visualizing high-dimensional data is an essential task in Data Science and Machine Learning. 
The  Centroid-Encoder (CE) method is similar to the autoencoder but incorporates label information to keep objects of a class close together in the reduced visualization space.
CE exploits nonlinearity and labels to encode  high variance in low dimensions 
while capturing the global structure of the data. 
% Centroid-encoder is simple to understand and easy to implement. 
We present a detailed analysis of the method using a wide variety of data sets and compare it with other supervised dimension reduction techniques, including NCA, nonlinear NCA, t-distributed NCA, t-distributed MCML, supervised UMAP, supervised PCA, Colored Maximum Variance Unfolding, supervised Isomap, Parametric Embedding, supervised Neighbor Retrieval Visualizer, and Multiple Relational Embedding. We empirically show that centroid-encoder outperforms most of these techniques.  We also show that when the data variance is spread across multiple modalities, centroid-encoder extracts a significant amount of information from the data in low dimensional space. This key feature establishes its value to use it as a tool for data visualization.
\end{abstract}

% keywords can be removed
\keywords{Data Science, Machine Learning, supervised dimension reduction, centroid-encoder, autoencoder, variance}

\section{Introduction}
Visualization of data is the process of mapping 
points in dimensions greater than three down to two or three
dimensions, providing a window into high-dimensions so that they can be {\it seen}
and interpreted.   The ability  to see in high-dimensions
can assist data explorers in many ways,
including the identification of experimental batch effects, 
data separability, and class structure.
Principal component analysis
\cite{pearson_1901,hotelling_1933}
continues to be
one of  the most widely used methods for data visualization  over one hundred years after its initial discovery \cite{jolliffe2016principal}.  Nonetheless,
PCA often produces ambiguous results,
in some cases collapsing distinct classes into overlapping regions 
in the setting where class labels are available.
It is tempting to incorrectly infer 
from this that the data is not separable, even nonlinearly,
in higher dimensions.

Nonlinear extensions to PCA were originally introduced to address the limitations of optimal linear
mappings  \cite{Kramer1,Kramer2,Oja91}, also see \cite{kirby_wiley2} for additional early references and details. These papers 
provided the first applications of autoencoder neural networks
where data sets are nonlinearly mapped to themselves. This is accomplished by first unfolding them
in higher dimensions before passing through a bottleneck layer
of a reduced dimension where data visualization is typically done.
In the process of nonlinear dimension reduction, 
novel {\it latent} or hidden features which 
are an amalgamation of the observables, may be discovered. A theoretical insight into these algorithms is 
provided by Whitney's theorem from differential geometry that connects
autoencoders to manifold learning \cite{kirby_1998a,kirby_1998b}.
Fundamental innovations were proposed to these ideas in the setting of deep networks that made 
their application far more effective \cite{hinton2006reducing,Hinton:2006:FLA:1161603.1161605}.

The development of tools for data visualization is predicated by the design of the objective function
and whether it incorporates class label information, i.e.,
is it supervised or unsupervised.  The objective function
encodes the goals of the dimension reduction,
i.e.,  what important information about the data is to
be captured.  This might be  geometric or topological structure (shape),
neighborhood relationships, or
descriptive statistics such as class distribution data
when data class labels are available.

%STAGE SET FOR DISCUSSING LABELS
This paper concerns the integration of label information to
the nonlinear autoencoder reduction process for the purposes of data visualization.
Here we develop and comparatively evaluate this {\it centroid-encoder} (CE) algorithm designed for the analysis of labeled data.   While standard 
autoencoders map the identity on points,
centroid-encoders map points to their 
centroids while passing through a low-dimensional representation.  This approach
serves to provide a low-dimensional  encoding for visualization while  ensuring that elements with the same label retain their class structure.  The smoothness of mapping functions ensures that a
similar behavior is captured at the centroid-encoder bottleneck layer.

In this paper, we analyze centroid-encoder and compare its performance with other methods. 
The article is organized as follows: 
In Section \ref{lit}, we review related work, both supervised and
unsupervised.  In Section \ref{AE}, we discuss the
autoencoder neural network for nonlinear
data reduction.  In Section \ref{CE}, we present
modifications to the autoencoder that 
result in centroid encoding.
In Section \ref{algorithm}, we present the 
training algorithm employed in this paper.
In Section \ref{experiments}, we illustrate the
application of CE to a range of bench-marking
data sets taken from the literature.
In Section \ref{discussion}, we analyze these results and conclude in Section \ref{conc}.

\section{Related Work}
\label{lit}
Data reduction for visualization has a long history and
remains an area of active research.   We 
describe the literature relating to unsupervised
and supervised methods where we assume
data class labels are 
unavailable and available, respectively.

\subsection{Unsupervised Methods}
 Principal Components Analysis is often the
first (and last) tool used for visualization of unlabeled data and is designed to 
retain as much of the statistical variance as possible \cite{pearson_1901,hotelling_1933}, see also  \cite{jolliffe_1986}.  Equivalently,
 PCA minimizes mean-square approximation error
 as well as Shannon's entropy \cite{watanabe_1965}. 
Self-organizing mappings (SOMs)
learn  nonlinear {\it topology-preserving} transformations
that map data points to centers that are then mapped
to the center indices.  SOMs have been widely 
used in data visualization, having been cited over
20,000 times since their introduction \cite{Kohonen_82}.

Another class of methods uses interpoint distances
as the starting point for data reduction and visualization.  For example,  Multidimensional Scaling  is a spectral method that computes a point configuration  based on the computation of the eigenvectors of a doubly centered distance matrix  \cite{Torgerson52}.   The goal of the optimization problem behind MDS is to determine a configuration 
of points whose Euclidean distance matrix is optimally close to the  prescribed distance matrix.  A related approach, known as Isomap, applies MDS to approximate 
geodesic distances computed numerically from data on a manifold \cite{Tenenbaum2319}.   Laplacian
eigenmaps is another popular  spectral method
that uses distance matrices to reduce dimension and conserve neighborhoods \cite{Belkin:2003:LED:795523.795528}.
These spectral methods belong to a class of techniques referred to as {\it manifold learning} algorithms;
see also locally linear embedding \cite{Roweis00nonlineardimensionality},
stochastic neighbor embedding \cite{NIPS2002_2276}, and maximum variance unfolding \cite{Weinberger:2006:IND:1597348.1597471}.

The technique t-distributed stochastic neighbor embedding \cite{vanDerMaaten2008}, an extension of SNE, was developed to
overcome the data crowding or clumping problem often observed with manifold
learning methods and is currently a popular method for data visualization. 
More recently,  the uniform manifold approximation and projection (UMAP)
algorithm has been proposed,  which uses Riemannian geometry and fuzzy simplicity sets to create a low-dimensional and locally uniformly distributed embedding of the data \cite{mcinnes2018umap}.
It has been reported that the algorithm has a better run time than t-SNE and offers 
compelling visualizations.

Note that these  spectral methods including  MDS, Laplacian Eigenmaps and UMAP compute
 embeddings based on solving an eigenvector problem
 requiring the entire data set. Such methods do not
 actually create mappings that can be applied to 
 reduce the dimension of new data points without 
 repeating the computation or resorting to the use of
 landmarks \cite{de2004sparse}.  This, in contrast to methods such as PCA, SOM
 and autoencoders that serve as mappings for streaming data.

\subsection{Supvervised Methods}
Fisher's linear discriminant analysis (LDA) reduces the dimension
of labeled data by simultaneously
optimizing class separation and within-class scatter \cite{fisher36lda,dudaHart1973}.  Both LDA and PCA 
are linear methods in that they construct optimal projection matrices, i.e., linear transformations for reducing the dimension of the data.

It is, in general, possible to add labels to unsupervised methods
to create their supervised analogues.
A heuristic-based supervised PCA model first selects important features by calculating correlation with the labels and then applies standard PCA of the chosen feature set \cite{doi:10.1198/016214505000000628}.
Another supervised PCA technique, proposed by Barshan et al. \cite{Barshan:2011:SPC:1950989.1951174}, uses a Hilbert-Schmidt independence criterion to compute the principal components which have maximum dependence on the labels. 
Colored maximum variance unfolding \cite{Song2007ColoredMV}, which is the supervised version of MVU, is capable of separating different new groups better than MVU and PCA.

The projection of neighborhood component analysis \cite{Goldberger:2004:NCA:2976040.2976105} produces a better coherent structure in two-dimensional space than PCA on several UCI data sets. Parametric embedding \cite{Iwata:2007:PEC:1288905.1288914}, which embeds high-dimensional data by preserving the class-posterior probabilities of objects, separates different categories of Japanese web pages better than MDS. Optimizing the NCA objective on a pre-trained deep architecture, Salakhutdinov et al. \cite{salakhutdinov2007learning} achieved $1\%$ error rate on MNIST test data with 3-nearest neighbor classifier on 30-dimensional feature space. Min et al. proposed to optimize the NCA and maximally collapsing metric learning \cite{globerson2006metric} objective using a Student t-distribution on a pre-trained network \cite{min2010deep}. Their approach yielded promising generalization error using a 5-nearest neighbor classifier on MNIST and USPS data on two-dimensional feature space.
Neighbor retrieval visualizer \cite{Venna:2010:IRP:1756006.1756019} optimizes its cost such that the similar objects are mapped close together in embedded space. Its supervised variant uses the class information to produce the low dimensional embedding. NeRV and its supervised counterpart were reported to outperform some of the state-of-the-art dimension reduction techniques on a variety of datasets.

%adaptation of unsupervised methods to supervised.

Supervised Isomap \cite{Cheng:2012:SIB:2426805.2426860}, which explicitly uses the class information to impose dissimilarity while configuring the neighborhood graph on input data, has a better visualization and classification performance than Isomap.  A supervised method using Local Linear Embedding has been proposed by \cite{Zhang:2009:ESL:1594411.1594551}. \cite{Raducanu:2012:SND:2142120.2142221} have suggested a 
	Laplacian eigenmap based supervised dimensionality reduction technique. Stuhlsatz et al. proposed a generalized discriminant analysis based on classical LDA \cite{stuhlsatz2012feature}, which is built on deep neural network architecture. Min et al. proposed a shallow supervised dimensionality reduction technique where the MCML objective is optimized based on some learned or precomputed exemplars \cite{min2017exemplar}. 
Although UMAP is originally presented as an unsupervised technique, its software package includes the option to build a supervised model \footnote{Manual for Supervised UMAP is located at \url {https://readthedocs.org/projects/umap-learn/downloads/pdf/latest/}}. UMAP is available in Python's Scikit-learn package \cite{Pedregosa:2011:SML:1953048.2078195}.

Autoencoder neural networks are the starting point for our approach and are described in the next section.   In a preliminary biological application,
centroid-encoder (CE)  was used to visualize the high-dimensional  pathway data \cite{GHOSH201826}. 

\section{Autoencoder Neural Networks}
\label{AE}
 
An autoencoder is a  dimension reducing mapping that has been 
optimized  to approximate  the identity 
on a set of training data \cite{AIC:AIC690370209,kirby_wiley2,Bengio:2006:GLT:2976456.2976476}. 
The mapping is modeled as the composition of a dimension reducing 
mapping $g$ followed by a dimension increasing reconstruction mapping $h$, i.e.,
$$f(x) = h(g(x))$$
where the {\it encoder} $g$ is represented
$$g: U \in \mathbb{R}^n \rightarrow V \in \mathbb{R}^m$$
and the {\it decoder} $h$ is represented
$$h: V \in \mathbb{R}^m \rightarrow U \in \mathbb{R}^n$$
The construction of $g$ and $h$, and hence $f$,  is 
accomplished by solving the unconstrained optimization problem
\begin{equation}
\begin{aligned}
 \underset {\theta} {\text{minimize}}\;\; \sum_i \| x^i - f(x^i; \theta)\|_2^2
 \label{equation:AEOptimization}
\end{aligned}
\end{equation}
for a given set of points $X=\{x^{1},x^{2},x^{3},...\}$. Hence, the autoencoder learns a function $f$ such that $f(x^i; \theta)\approx x^i$ where $\theta$ is the set of model parameters that is found by solving the optimization problem. The encoder map $g$ takes the input $x^i$ and maps it to a latent representation 
$$y^i = g(x^i) \in V \subset \mathbb{R}^m$$ 
In this paper we are primarily concerned with visualization so we take $m=2$.
The decoder ensures that the encoder is faithful to the data point, i.e., it
serves to reconstruct the reduced point to its original state
$$x^i \approx h(y^i)$$

The parameters $\theta$ are learned by using error backpropagation \cite{rumelhart1985learning,Werbos_thesis}.   The details for using multi-layer perceptrons for training autoencoders can be found in \cite{kirby_wiley2}.
 
If we restrict $g$ and $h$ to be linear transformations then \cite{Baldi:1989:NNP:70359.70362} showed that autoencoder function learns the same subspace of PCA. The nonlinear autoencoder has been proposed as a nonlinear version of PCA 
\cite{Kramer1,Kramer2}. If we restrict  $g$ to be linear and allow $h$ to be nonlinear then the autoencoder
fits into the blueprint of Whitney's theorem for manifold embeddings \cite{kirby_1998a,kirby_1998b}. Topologically sensitive nonlinear reduction use, e.g., topological  circle or sphere encoding neurons \cite{kirby_1996sub1,kirby_1996,kirby_1995a} to capture shape in data. The use of shallow architectures was partly motivated by the theoretical 
result that a neural network with one hidden layer is a universal approximator by \cite{Hornik:1989:MFN:70405.70408,Cybenko1989}.

A probabilistic pre-training approach, known as Restricted Boltzmann Machine (RBM), was introduced \cite{Hinton:2006:FLA:1161603.1161605,hinton2006reducing}, which initializes the network parameters near to a good solution followed by a nonlinear refinement to further fine-tune the parameters. This break-through opened the window of training neural networks with many hidden layers. Since then {\it deep neural networks} (DNN) architecture has been employed to train autoencoders.  The pre-training approach was extended further to continuous values by \cite{Bengio:2006:GLT:2976456.2976476,NIPS2006_3048}. Several other deep autoencoder based model were also proposed to extract sparse features \cite{Ranzato:2007:SFL:2981562.2981711,NIPS2007_3313}. The method of denoising autoencoder \cite{Vincent:2010:SDA:1756006.1953039} was introduced to learn latent features by reconstructing an input from it noisy version. Autoencoders can also be trained by stacking multiple convolutional layers as described by \cite{Masci:2011:SCA:2029556.2029563,DCCAE,DBLP:journals/corr/abs-1711-08763}. A comprehensive description of autoencoders and their variants can be found in \cite[chap.~14]{Goodfellow-et-al-2016}.

\section{The Centroid-Encoder}
\label{CE}
The autoencoder neural network does not employ any label information
related to the classes of the data. Here we propose a form of supervised autoencoder
that exploits label information.  The  centroid-encoder (CE) is trained by mapping data
through a neural network with the architecture of an autoencoder, but the 
learning target of
any point is not that actual point, rather the mean of points in the associated class.  
So CE does not map the identity, now the target points in the image of the map are 
the centroids of the data in the ambient space.
Centroid-encoder is a nonlinear parametric map ($f_{\theta}$ where $\theta$ is the set parameters) which minimizes the within-class variance by mapping all the samples of a class/category to its centroid. 
\subsection{Centroid-Encoder Backpropagation}
Consider an $M$-class  data set with classes denoted $C_j, j = 1, \dots, M$
where the indices of the data associated with class $C_j$ are denoted
$I_j$.  We define centroid of each class as
 $$c_j=\frac{1}{|C_j|}\sum_{i \in I_j} x^i$$ 
  $|C_j|$ is the cardinality of  class $C_j$.  The centroid-encoder
  is trained by determining a mapping $f = h \circ g$ with input-output pairs
$$\{ x^i, c_j\}_{i=1}^N$$
where $c_j$ is the target center for data point $x^i$, i.e., $i \in I_j$.
%For a labeled data set with $N$ input-output pairs $\{x^i,y^i\}^N_{i=1}$, where $x^i \in {\mathbb{R}}^d$ and $y^i$ is the class label, 
Thus, unlike autoencoder, which maps each point $x^i$ to itself, centroid-encoder will map each point $x^i$  to its class centroid $c_j$ 
$$c_j \approx f(x^i)$$
while passing the data through a bottleneck layer of dimension $m$ to 
provide the reduced point $g(x^i)$.  
The cost function of centroid-encoder is defined as
\begin{equation}
\begin{aligned}
 \mathcal{L}_{ce}(\theta)=\frac{1}{2N}\sum^M_{j=1} \sum_{i \in I_j}\|c_j-f(x^i; \theta))\|^2_2
 \label{equation:CECostFunction}
\end{aligned}
\end{equation}
This cost is also referred to as the {\it distortion error} (\cite{L-B-G}).
The network parameters ($\theta$) are trained using the  error backpropagation
appropriately modified for the centroid targets. 
 We can connect this to the learning procedure for autoencoders by
 first writing
 $$\mathcal{L}_{ce}(\theta) = E(f(\theta))$$
 where the $f(\theta)$ is identical for 
 autoencoders and centroid-encoders.
 Then, applying the chain rule,  we have
$$\frac{ \partial{\mathcal{L}_{ce}(\theta)} }{\partial \theta_i} = 
 \frac{\partial E}{\partial f}
\frac{\partial f}{\partial \theta_i}$$
where the term
$\frac{\partial f}{\partial \theta_i}$ is the same as for autoencoders
and $\theta _i$ is the $i$th component of the parameter vector $\theta$.
The new term $ \frac{\partial E}{\partial f}$ is readily calculated to be

\begin{equation}
\frac{\partial E}{\partial f} = \frac{1}{N}\sum^M_{j=1} \sum_{i \in I_j} (f(x^i,\theta) - c_j)
\label{equation:OutputLayerDeltas}
\end{equation}

\section{The Training Algorithm}
\label{algorithm}
%The map $f_{en}$, which is known as encoder, takes a $d$ dimensional input $x_i\in C_j$ and maps it to a hidden representation $h_i$ where $h_i \in {\mathbb{R}}^m$ and $m<d$. The map $f_{de}$, which is known as decoder, takes the latent code $h_i$ and maps it to $\bar{C_j}$. 
The training is very similar to that of an autoencoder and the steps are described in Algorithm \ref{algorithm:centroid-encoder}.
As with autoencoders,  the centroid-encoder is a composition of two maps $f(x)=(h \circ g)(x)$. 
For visualization, we train a centroid-encoder network using a bottleneck architecture,
meaning that the dimension of the image 
of the map $g$ is 2 or 3.

\RestyleAlgo{boxruled}
\LinesNumbered
\begin{algorithm}[!ht]

\SetKwInput{KwData}{Input Data}
\KwData{Labeled Data ${\{x^i\}}^N_{i=1}\, \in {\mathbb R}^n,\, \, I_j$ index set of class $C_j.$  User defined parameters training epoch $n\_epochs$ and visualization dimension $m$.}
 \SetKwInput{KwData}{Output Data}
 \KwData{Bottleneck layer $y^i = g(x^i)$; Output layer $h(y^i), i = 1, \dots, N.$ }
\SetKwInput{KwData}{Nonlinear function}
 \KwData{$f: {\mathbb R}^n \to {\mathbb R}^n $ where $f = h \circ g $.}
 \KwResult{Nonlinear embedding of data in $m$ dimensions.}
 \SetKwInput{KwData}{Initialization}
\KwData{ 
Calculate the class centroids $c_j = \frac{1}{|C_j|} \sum_{i \in I_j}^{}{x^i}, j = 1,\dots, M$.}
  \SetKwInput{KwData}{Define}
\KwData{Centroid-encoder output error $E = \frac{1}{2N}\sum_{j} \sum_{i \in I_j} \|c_j-f(x^i,\theta)\|^2_2$}

 \For{$k=1$ to $n\_epochs$}{
  
  Update parameters $\theta$ of centroid-encoder. \
 }
 \caption{Supervised Nonlinear Centroid-Encoder.}
 \label{algorithm:centroid-encoder}
\end{algorithm}

\subsection{Pre-training Centroid-encoder}
\begin{figure}[!ht]
        %\centering        
        \includegraphics[width=\textwidth]{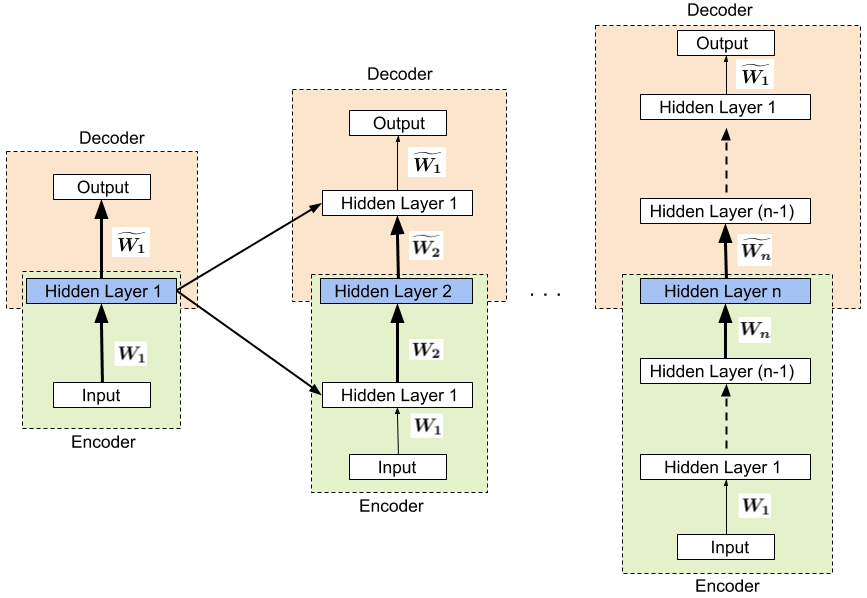}
        \caption{Pre-training a deep centroid-encoder by layer-freeze approach. In the first step, a centroid-encoder with the first hidden layer is pre-trained (left diagram). The pre-trained weights are ($W_1,\widetilde{W_1}$). In the next step, a new hidden layer is added by extending the network architecture. After that, the weights ($W_2,\widetilde{W_2}$) associated with the new hidden layer are updated while keeping the other weights ($W_1,\widetilde{W_1}$) fixed. This process is repeated to add more hidden layers.}\label{fig:CEPreTraining}
\end{figure}

Here we describe the pre-training strategy of centroid-encoder.  We employ a technique we
refer to as 
{\it pre-training with layer-freeze}, i.e.,  pre-training is done by adding new hidden layers while mapping samples of a class to its centroid;   see Figure \ref{fig:CEPreTraining}. In this approach, weights of hidden layers are learned sequentially. The algorithm starts by learning the parameters of the first hidden layer using standard backpropagation. Then the second hidden layer is added in the network with weights initialized randomly. At this point, the associated parameters of the first hidden layer are kept frozen. Now the weights of the second hidden layer are updated using backpropagation. We repeat this step until pre-training is done for each hidden layer. Once pre-training is complete, we do an end-to-end fine-tuning by updating the parameters of all the layers at the same time. It's noteworthy that our pre-training approach is different than the greedy layer-wise pre-training proposed by \cite{Hinton:2006:FLA:1161603.1161605} and \cite{Bengio:2006:GLT:2976456.2976476,NIPS2006_3048}.  The unsupervised pre-training of Hinton et al. and Bengio et al. uses the activation of the $l^{th}$ hidden layer as the input for the $(l+1)^{th}$ layer. In our approach, we used the original input-output pair $(x^i,c_j)$ to pre-train each hidden layer. As we use the labels to calculate the centroids $c_j$, so our pre-training is supervised. In the future, we would like to explore the unsupervised pre-training in the setup of centroid-encoder.

\section{Classification and Visualization Experiments}
\label{experiments}

We select three suites of data sets from the literature 
with which we conduct three bench-marking experiments comparing  centroid-encoder with a range of other  supervised dimension reduction techniques.   
We describe the details of the data sets in Section \ref{dataset}. In Section \ref{methodology}, we describe the comparison methodology  and present the results of the experiments in Section \ref{quantviz}.

\textbf{Experiment 1:}  The first bench-marking experiment  is conducted on the
widely studied MNIST and USPS data sets (see Section \ref{dataset} for details).   In this experiment we compare CE with the following methods: autoencoder, nonlinear NCA \cite{salakhutdinov2007learning}, supervised UMAP \cite{mcinnes2018umap}, GerDA \cite{stuhlsatz2012feature}, HOPE \cite{min2017exemplar}, parametric t-SNE \cite{DBLP:journals/jmlr/Maaten09}, t-distributed NCA \cite{min2010deep} and t-distributed MCML \cite{min2010deep}, and supervised UMAP. Parametric t-SNE is a DNN based model with RBM pre-training, where the objective of t-SNE is used to update the model parameters (hence parametric t-SNE). Although it is an unsupervised method, we included it as t-SNE has become a baseline model to compare visualization. We also included autoencoder as it is widely used for data visualization. We implemented nonlinear-NCA in Python, and we used the scikit-learn \cite{Pedregosa:2011:SML:1953048.2078195} package to run supervised UMAP. For the rest of the methods, we took the published results for comparison.

\textbf{Experiment 2:} We conducted the second experiment on Phoneme, Letter, and Landsat data sets (see Section \ref{dataset} for details). In this case, we compared CE with the following techniques: supervised neighbor retrieval visualizer (SNeRV) by \cite{Venna:2010:IRP:1756006.1756019}, multiple relational embedding (MRE) by \cite{Memisevic:2004:MRE:2976040.2976155}, colored maximum variance unfolding (MUHSIC) by \cite{Song2007ColoredMV}, supervised isomap (S-Isomap) by 
\cite{Cheng:2012:SIB:2426805.2426860}, parametric embedding (PE) by \cite{Iwata:2007:PEC:1288905.1288914} and neighborhood component analysis (NCA) by \cite{Goldberger:2004:NCA:2976040.2976105}. We didn't implement these models; instead, we used the published results from \cite{Venna:2010:IRP:1756006.1756019}. On the Landsat data set, we ran UMAP and SUMAP along with the methods as mentioned above.

\textbf{Experiment 3:} For the third experiment, we compared the performance of centroid-encoder with supervised PCA (SPCA), and kernel supervised PCA (KSPCA) on the Iris, Sonar, and a subset of USPS dataset (details of these dataset are given in Section \ref{dataset}). We implemented SPCA and KSPCA in Python following \cite{Barshan:2011:SPC:1950989.1951174}.

\subsection{Data Sets}
\label{dataset}
Here we provide a  brief description of the data sets used in the bench-marking experiments.

\textbf{MNIST Digits:} This is widely used  collection of digital images of handwritten digits (0..9)\footnote{The data set is available at \url{http://yann.lecun.com/exdb/mnist/index.html}.} with separate training (60,000 samples) and test set (10,000 samples).   Each sample is a  grey level  image consisting of 1-byte pixels normalized to fit into a 28 x 28 bounding box resulting  in {\it vecced}  points in $\mathbb{R}^{784}$.

\textbf{USPS Data:}  A data set of handwritten digits ($0\dots 9$)\footnote{The data set is available at \url{https://cs.nyu.edu/~roweis/data.html}.}, where each element is a 16x16 image in gray-scale resulting in
{\it vecced} points in $\mathbb{R}^{256}$. Each of the ten classes has 1100 digits for a  total of  11,000 digits. 

\textbf{Phoneme Data:} This data set is created from Finish speech recorded continuously from the same speaker. A 20-dimensional vector represents each phoneme. The data set consists of 3924 number of samples distributed over 13 classes. The data set is taken from the LVQ-PAK\footnote{The software is avilable at \url{http://www.cis.hut.fi/research/software}.} software package.

\textbf{Letter Recognition Data:} Available in the UCI  Machine Learning Repository \cite{Dua:2019}, this data set is comprised of  20,000 samples where each element  represents an upper case letter of the English alphabet A to Z selected from twenty different fonts.  Each sample is randomly distorted to produce a unique representation that is converted to 16 numerical values and 26 classes.

\textbf{Landsat Satellite Data:} This data set consists of satellite images with four different spectral bands. Given the pixel values of a 3x3 neighborhood, the task is to predict the central pixel of each 3x3 region. Each sample consists of 9-pixel values from 4 different bands, which makes the dimension 36.  There are six different classes and 6435 samples. The data set is also available in UCI Machine Learning Repository \cite{Dua:2019}.

\textbf{Iris Data:} The UCI Iris data  set \cite{Dua:2019} is comprised of three classes and is widely used in the  Machine Learning literature. Each class has 50 samples and, each sample has four features.

\textbf{Sonar Data:} This UCI data set \cite{Dua:2019} has two classes: mine and rock. Each sample is represented by a 60 dimensional vector. The mine class has 111 patterns, whereas the rock category has 97 samples.

In the next section we describe the methodology used to compare the low-dimensional embedding of CE with other methods on the data sets described above.

\subsection{Methodology}
\label{methodology}
To objectively compare  the performance of CE  with other techniques in the literature, we employ
a standard class prediction error
on the two-dimensional visualization domain.\footnote{This corresponds to the bottleneck layer for CE.}
As we are comparing supervised methods, we restrict our attention to the generalization performance as
measured by a test data set.  

The  class prediction error
 is defined as 
\begin{equation}
\begin{aligned}
 Error\;(\%) =\frac{100}{N}\sum^N_{i=1} I[l_i \neq f(\tilde{x_i})]
 \label{equation:embeddingError}
\end{aligned}
\end{equation}
 Here $N$ is total number of test samples, $l_i$ is the true label of the $i^{th}$ test sample,  and $f$ is a  classification function which returns the predicted label of the embedded test sample $\tilde{x_i}$. Here $I$ denotes the indicator function, which has the value 1 if the argument is true (label incorrect), and 0 otherwise (label correct).

A common choice of $f$ for bench-marking 
algorithms in the supervised visualization literature 
is the $k$-nearest neighbor (k-NN) classifier \cite{Goldberger:2004:NCA:2976040.2976105,Venna:2010:IRP:1756006.1756019,DBLP:journals/jmlr/Maaten09,min2010deep}. Thus,  the evaluation proceeds as follows: for each method divide the data into training and testing sets;  train the model on the training set and then map  the test set through the trained model to get the low dimensional (2D) representation. Finally,  classify 
each test point in the 2D visualization space by k-NN algorithm. 
The prediction error indicates the percentage  of
wrongly classified samples on the test data and is an objective  measure of the quality of the visualization. A low error rate suggests that the samples from the same class are close together in the visualization  space.

Like autoencoder, centroid-encoder is a model based on deep architecture whose performance varies based on network topology.  In addition to network architecture, both the models require the following hyper-parameters: learning rate\footnote{Learning rate was selected from the following list of values: 0.1, 0.01, 0.001, 0.0001, 0.0002, 0.0004, 0.0008.}, mini-batch\footnote{Mini-batch size was selected from the following values: 16, 32, 50, 64, 128, 256, 512, 1024.} size, and weight decay\footnote{Weight decay was chosen from the following values: 0.001, 0.0001, 0.00001, 0.00002, 0.00004, 0.00008.} constant. As with all neural network learning algorithms, these parameters need to be tuned to yield optimal  performance. 
The details of the training procedure vary by data set.
We used subsets of the 
training data (3000 and 30,000 for USPS and MNIST respectively), picked randomly, and ran 10-fold cross-validation to determine the network architecture and hyper-parameters.  On MNIST, we trained CE on the entire training set and then used the test data to calculate class prediction error using a $k$-NN ($k=5$) algorithm. On USPS data, we followed the strategy of \cite{min2010deep}, where we randomly split the entire data set into a training set of 8000 samples and a test set consisting of 3000 samples. After building the model on training samples, we used the test set to calculate the error using $k$-NN with $k$=5. In both cases, we repeat the experiment 10 times and report the average error rate with standard deviation.

The supervised UMAP (S-UMAP) and nonlinear NCA (NNCA), also require the tuning of hyper-parameters. For NNCA, we used the architecture input dimension $d \rightarrow500\rightarrow500\rightarrow2000\rightarrow 2$ as proposed by \cite{min2010deep};   we also followed the same pre-training and fine-tuning steps described there. 
The algorithm S-UMAP requires two hyper-parameters including, the  number of neighbors ($n\_neighbors$\footnote{$n\_neighbors$ was picked from the following list of values: 5, 10, 20, 40, 80.}) and minimum distance ($min\_dist$\footnote{$min\_dist$ was picked from the following list of values: 0.0125, 0.05, 0.2, 0.8.}) allowed between the points in reduced space. On the MNIST and USPS, we used 10-fold cross-validation on a randomly-selected subset of training data (3000 and 30,000 for USPS and MNIST respectively) to optimize these parameters.

On Phoneme, Letter, and Landsat data, our experimental setup is similar to that of \cite{Venna:2010:IRP:1756006.1756019}. We randomly selected 1500 samples from each dataset and ran 10-fold cross-validation.
We picked the hyper-parameters (network architecture, learning rate, mini-batch size, weight decay) by running 10-fold internal cross-validation on the training set.  After choosing the model parameters, we trained the model with the full training data and calculated the error rate using $k$-NN ($k$=5) on the holdout test set. Once all the 10 test sets are evaluated, we reported the average test error with standard deviation.

\begin{table}[!ht]	
	\vspace{1mm}
	\centering
	\begin{tabular} {|c|c|c|c|c|}	% (4 columns)
		% start header
		\hline\hline 	% Makes 2 fancy lines
		Dataset & Dimensionality & No. of Classes & No. of Training & No. of Test \\
		& & & Samples & Samples \\
		\hline
		MNIST & 784 & 10 & 60,000 & 10,000 \\
		USPS & 256 & 10 & 8000 & 3000 \\
		Phoneme & 20 & 13 & 1,350 & 150 \\
		Letter & 16 & 26 & 1,350 & 150 \\
		Landsat & 36 & 6 & 1,350 & 150 \\
		Iris & 4 & 3 & 105 & 45\\
		Sonar & 60 & 2 & 146 & 62\\
		USPS 1000 & 256 & 10 & 700 & 300 \\
		\hline \hline
	\end{tabular}	
	\caption{Details of data sets along with the sample count of training and test set used in the visualization experiments.}
	\label{table:DataSetDetails}
\end{table}

We followed the experimental setup of \cite{Barshan:2011:SPC:1950989.1951174} in the last experiment, where we compared centroid-encoder with supervised PCA (SPCA) and kernel supervised PCA (KSPCA). Each of the three data sets (Iris, Sonar, USPS) is split randomly into training and test by the ratio of 70:30. After training the models, the generalization error is calculated on the test set using $k$-NN ($k$=5). We repeat the experiment 25 times and report the average error with the standard deviation for all three data sets. For the USPS set, we randomly picked 1000 cases and used that subset for our experiment, as done by \cite{Barshan:2011:SPC:1950989.1951174}. Among the three models, centroid-encoder and KSPCA require hyper-parameters. For KSPCA we apply RBF kernel on data and delta kernel on labels, as recommended by Barshan et al. The delta kernel doesn't require any parameter, whereas the RBF kernel needs the parameter $\gamma$ (width of the Gaussian\footnote{$\gamma$ was picked from the following values: 0.001, 0.01, 0.025, 0.035, 0.05, 0.1, 1.0, 2.5, 3.0, 5.0, 7.5, 10.0.}) which was picked by running 10-fold internal cross-validation on the training set. Table \ref{table:DataSetDetails} gives the sample count of training and test sets of each data set along with other relevant information. We used two versions of USPS data: in Experiment 1, we used the entire data set, and in Experiment 3, we used a subset (USPS 1000).

\begin{table}[!ht]	
	\vspace{1mm}
	\centering
	\begin{tabular} {|c|c|c|c|c|c|}	% (3 columns)
		% start header
		\hline\hline 	% Makes 2 fancy lines
		Dataset & Network topology & Activation & Learning & Batch & Weight \\
		&  & function & rate & size & Decay\\
		\hline
		MNIST & $d \rightarrow1000\rightarrow500\rightarrow125\rightarrow 2$ & tanh & 0.0008 & 512 & 2e-5\\
		%\hline
		USPS & $ d\rightarrow2000\rightarrow1000\rightarrow500\rightarrow 2$ & relu & 0.001 & 64 & 2e-5\\
		Phoneme & $d \rightarrow250\rightarrow150\rightarrow 2$ & relu & 0.01 & 50 & 2e-5\\
		Letter & $d\rightarrow250\rightarrow150\rightarrow 2$ & relu & 0.01 & 50 & 5e-5\\
		Landsat & $d \rightarrow250\rightarrow150\rightarrow 2$ & relu & 0.01 & 50 & 5e-5\\
		Iris & $d \rightarrow100\rightarrow 2$ & relu & 0.001 & 16 & 2e-5\\
		Sonar & $d \rightarrow500\rightarrow 250\rightarrow 2$ & relu & 0.001 & 16 & 2e-5\\
		\hline \hline
	\end{tabular}	
	\caption{Details of network topology and hyper-parameters for CE. The number $d$ is the input dimension of the network and
		is data set dependent.}
	\label{table:CEHyperparameters}
\end{table}

To train CE with an optimal number of epochs, we used $10\%$ of training samples as a validation set in all of our visualization experiments. We measure the generalization error on the validation set after every training epochs. Once the validation error doesn't improve, we stop the training. Finally, we merge the validation set with the training samples and train the model for additional epochs. The model parameters are updated using Adam optimizer (\cite{DBLP:journals/corr/KingmaB14}). Table \ref{table:CEHyperparameters} gives the details of network topology and hyper-parameters used in training of CE. We have implemented centroid-encoder in PyTorch to run on GPUs. The implementation  is available at: \url{https://github.com/Tomojit1/Centroid-encoder/tree/master/GPU}.

In addition to computing the prediction error, we also visualize the two-dimensional embedding of test samples of each model. For centroid-encoder, we plot the bottleneck output of training and test data using Voronoi regions.  Here, the bottleneck outputs of training data are used to compute the centroids of each class, which are used to create the Voronoi cells.   The test set was then mapped to the reduced space. This visualization approach allows one to see the quality of the CE classification directly.

\subsection{Quantitative and Visual Analysis}
\label{quantviz}
Now we present the results from a comprehensive quantitative and qualitative analysis
across diverse data sets.  Note that the primary objective of our assessment is 
to determine the quality of information obtained from visualization, including class separability, data scatter, 
and neighborhood structure.   Since this assessment is by its very nature subjective, we also employ label information to determine classification rates that provide additional insight into the data reduction
for visualization.  Computational expense is also an essential factor, and we will see that this
is a primary advantage of CE over other methods when the visualizations 
and error rates are comparable.

\subsubsection{MNIST}

As shown in Tables \ref{table:MNIST_USPS_Result_W_pretr} (results with pre-training) 
and  \ref{table:MNIST_USPS_Result_WO_pretr} (results without  pre-training),
the models with relatively low error rates for the MNIST data amongst our suite of visualization methods are dt-MCML, dG-MCML, centroid-encoder, and HOPE.   
The error rate of CE is comparable to the top-performing model dt-MCML
and superior to NNCA and supervised UMAP by a margin of $2.1 \%$ and $0.84 \%$, respectively. { Note that methods in the MCML and NCA families require the computation of 
distance matrix over the data set, making them significantly more expensive 
than CE, which only requires distance computations between the data of a class and its center.
}
It's noteworthy that the error of CE is lower than the other non-linear variants of NCA: dt-NCA and dG-NCA. Among all the models, parametric t-SNE (pt-SNE) and autoencoder exhibit
 the worst performance  with error rates on MNIST data are $9.90\%$ and $22.35\%$, respectively.  These high error rates are not surprising given these methods   do not use label information.  

The prediction error of CE with pre-training is relatively low at $2.6\%$, as shown in Table \ref{table:MNIST_USPS_Result_W_pretr} behind the more computationally expensive dt-MCML and  dG-MCML algorithms. 
 With no pre-training,  the numeric performance of centroid-encoder is statistically equivalent to dt-MCML and HOPE.
 
\begin{table}[!ht]
	\centering
	\begin{tabular} {|c|c|c|}	% (3 columns)
		% start header
		\hline\hline 	% Makes 2 fancy lines
		\multirow{2}{*}{Method} & \multicolumn{2}{c|} {Dataset}\\
		\cline{2-3}		
		& \multicolumn{1}{c|} {MNIST} &  \multicolumn{1}{c|} {USPS} \\
		
		\hline
		Centroidencoder & $2.61 \pm 0.09$ & $2.91 \pm 0.31$ \\
		Supervised UMAP & $3.45\pm 0.03$ & $6.17\pm 0.23$\\
		NNCA & $4.71\pm 0.57$ & $6.58\pm 0.80$\\
		Autoencoder & $22.04 \pm 0.78$ & $16.49 \pm .91$ \\
		\hline		
		dt-MCML & $\textbf{2.03}$ & $\textbf{2.46}\pm \textbf{0.35}$\\
		dG-MCML & $2.13$ & $3.37\pm 0.18$\\		
		GerDA & $3.2$ & NA\\
		dt-NCA & $3.48$ & $5.11\pm 0.28$\\
		dG-NCA & $7.95$ & $10.22\pm 0.76$\\
		pt-SNE & $9.90$ & NA\\
		\hline \hline
	\end{tabular}	
	\caption{Error rates ($\%$) of $k$-NN ($k$=5) on the 2D embedded data by various dimensionality reduction techniques trained with pre-training. Results of GerDA and pt-SNE are taken from \cite{stuhlsatz2012feature} and \cite{DBLP:journals/jmlr/Maaten09} correspondingly. Error rates of the variants of NCA and MCML are reported from \cite{min2010deep}. NA means result is not available.}
	\label{table:MNIST_USPS_Result_W_pretr}
\end{table}

\begin{table}[ht!]
	\centering
	\begin{tabular} {|c|c|c|}	% (3 columns)
		% start header
		\hline\hline 	% Makes 2 fancy lines
		\multirow{2}{*}{Method} & \multicolumn{2}{c|} {Dataset}\\
		\cline{2-3}		
		& \multicolumn{1}{c|} {MNIST} &  \multicolumn{1}{c|} {USPS} \\
		
		\hline
		Centroidencoder & $\textbf{3.17} \pm \textbf{0.24}$ & $\textbf{2.98} \pm \textbf{0.67}$ \\
		Autoencoder & $21.55\pm 0.47$ & $15.17\pm 0.85$\\
		\hline
		HOPE & $3.20$ & $3.03$\\		
		dt-MCML & $3.35$ & $4.07$\\		
		dt-NCA & $3.48$ & $5.11$\\
		\hline \hline
	\end{tabular}	
	\caption{Error rates ($\%$) of $k$-NN ($k$=5) on the 2-dimensional data by various techniques trained without pre-training. Error rate of HOPE, dt-NCA and dt-MCML are reported from \cite{min2017exemplar}.}
	\label{table:MNIST_USPS_Result_WO_pretr}
\end{table} 

\begin{figure}[pt!]
	\begin{subfigure}[b]{1.00\textwidth}
		\centering
		\includegraphics[width=12.0cm, height=9.65cm]{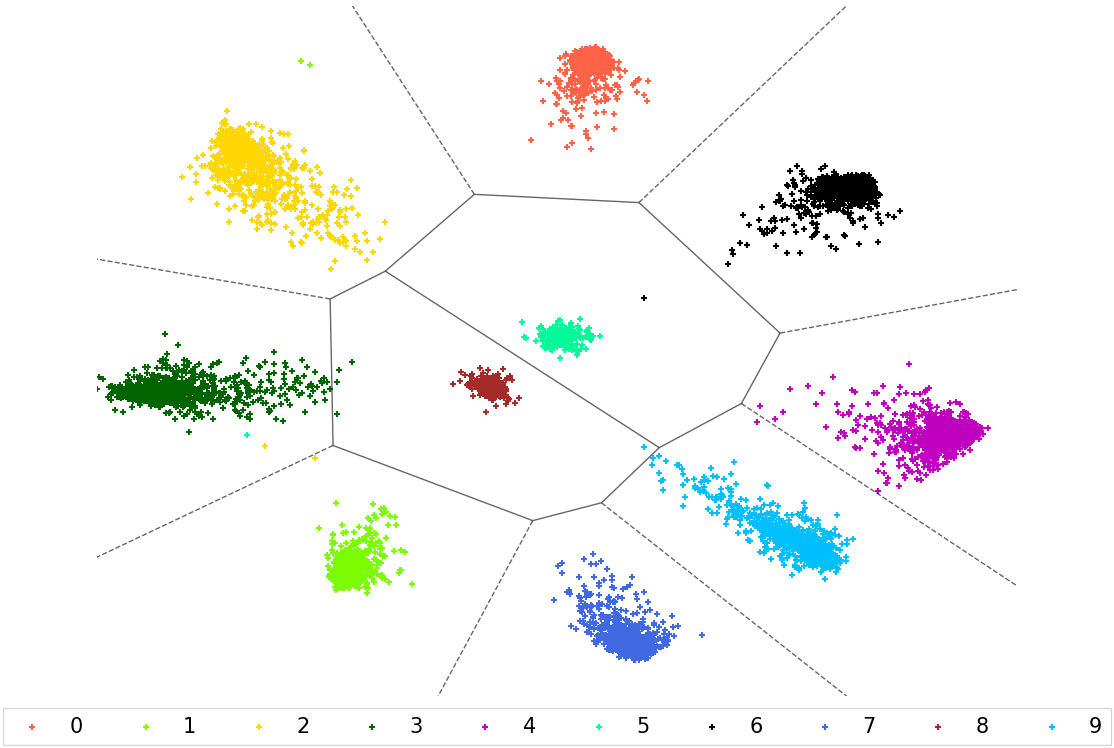}        
		\caption{Visualuzation of 1000 digits per class from MNIST training set.}
		\vspace{0.55cm}
	\end{subfigure}    
	\begin{subfigure}[b]{1.00\textwidth}
		\centering
		\includegraphics[width=12.0cm, height=9.65cm]{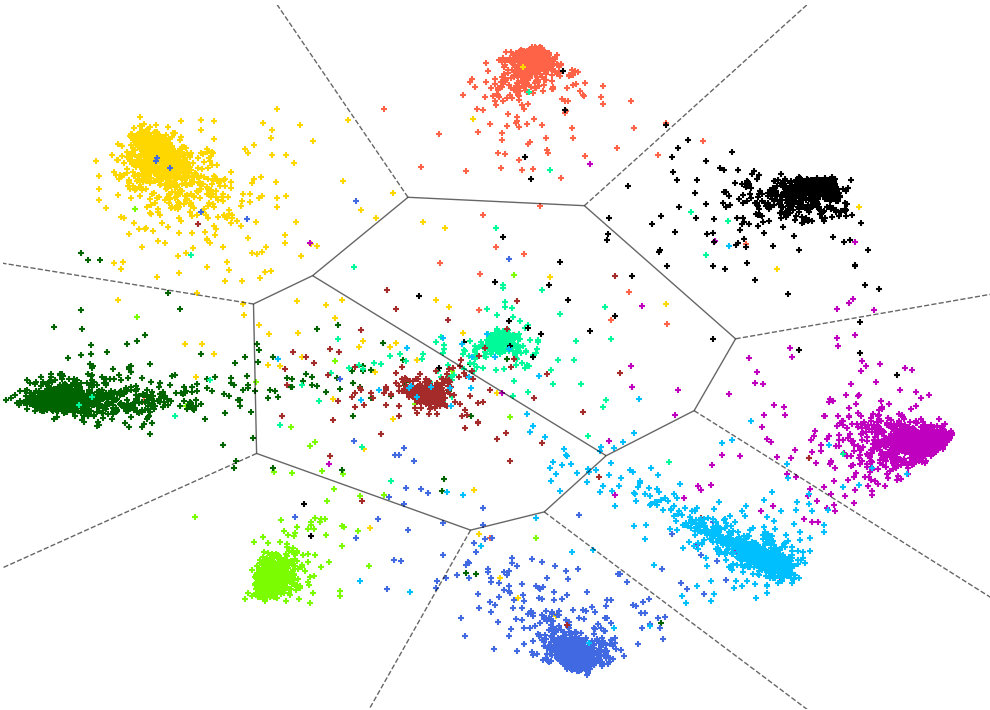}
		\caption{Visualization of the 10,000 MNIST test samples.}  
	\end{subfigure}   
	\caption{Voronoi cells in 2D of the MNIST data using centroid-encoder. The network architecture of $784\rightarrow[1000,500,125,2,125,500,1000]\rightarrow784$ with linear activation in the bottleneck layer is employed to map the data onto the 2D space. The centroid of the training samples mapped to 2D are used to form the Voronoi regions for each digit class.}
	\label{fig:2DCEofMNIST}
\end{figure}

While the prediction error is essential as a quantitative measure, 
the visualizations reveal information not encapsulated in this number.
The neighborhood relationships established by the embedding provide 
potentially valuable insight into the structure of the data set.
The two-dimensional centroid-encoder visualization of the MNIST  training  and test sets are shown in Figure \ref{fig:2DCEofMNIST}. The entire 10,000 test samples are shown in b) while only a subset of the training data set, 1000 digits picked randomly from each class, is shown in a). The separation among the ten classes is easily visible in both training and test data, although there are few overlaps among the categories in test data. Consistent with the low error rate, the  majority of the test digits are assigned to the correct Voronoi cells.

\newpage

\begin{figure}[ht!]
	\includegraphics[width=15.2cm, height=12.0cm]{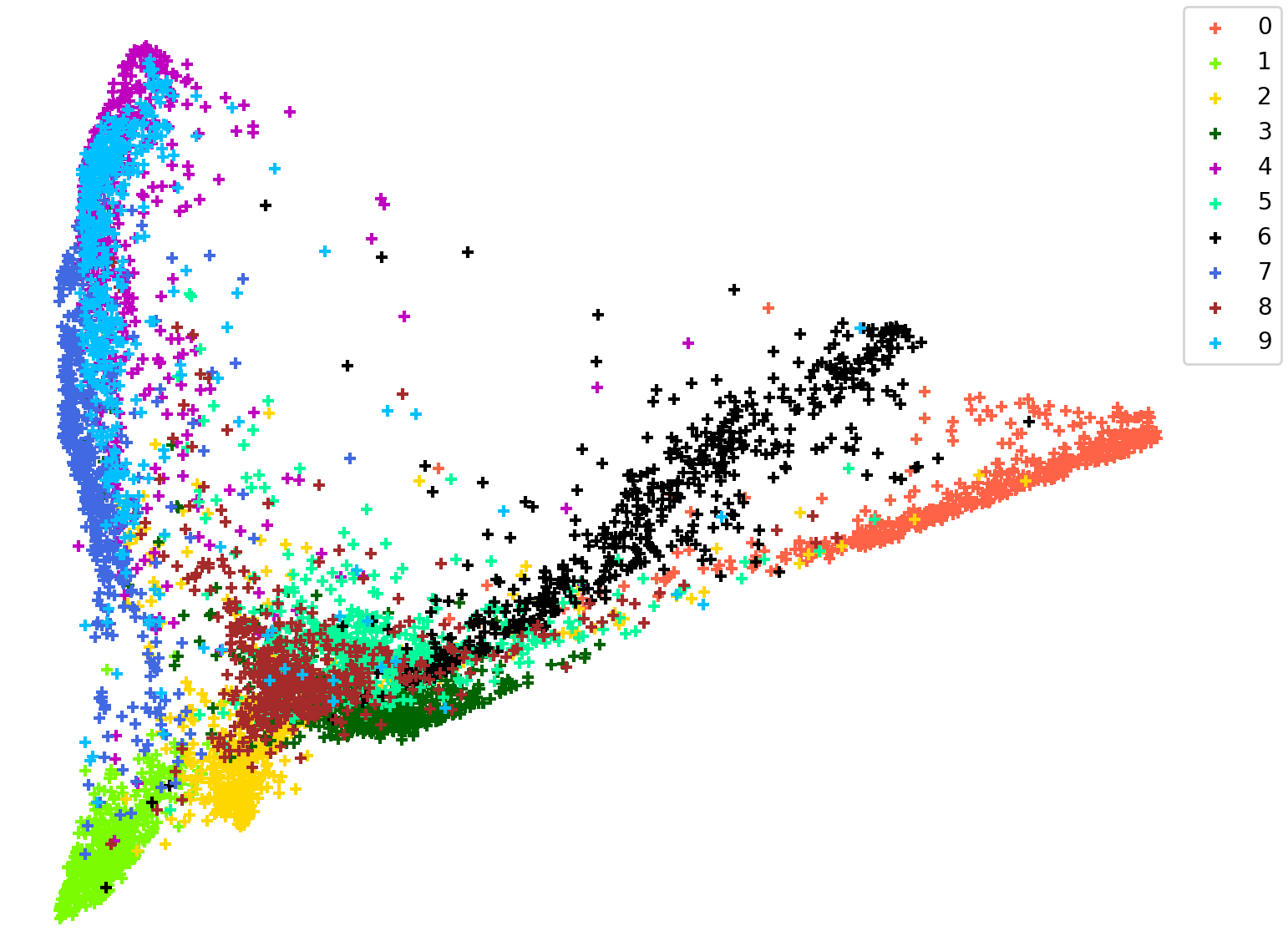}
	\caption{Two dimensional visualization  of  10000 MNIST test digits by Laplacian Eigenmap.}
	\label{fig:MNIST_2d_Tstdata_LE}
\end{figure}

Now let's consider the visualization from a neighborhood relationship perspective.  We compare the results with those of Laplacian Eigenmaps (LE), as shown in Figure \ref{fig:MNIST_2d_Tstdata_LE}, an unsupervised manifold learning spectral method that solves an optimization problem preserving neighborhood relations \cite{Belkin:2003:LED:795523.795528}. Both CE and LE place digits 5 and 8 as neighbors at the center,
allowing us to view all digits as being perturbations of these numbers. Digits 7, 9, and 4 are collapsed to one region; in contrast,  CE is built to exploit label information and separates these 
neighboring digits.  LE clumps 7, 9, 4 next to 1 as well, consistent with the CE visualization. The digit 0 neighbors 6 for both CE and LE, but LE significantly overlaps the 6 with  the digit 5.

\begin{figure}[ht!]
	\includegraphics[width=15.2cm, height=12.0cm]{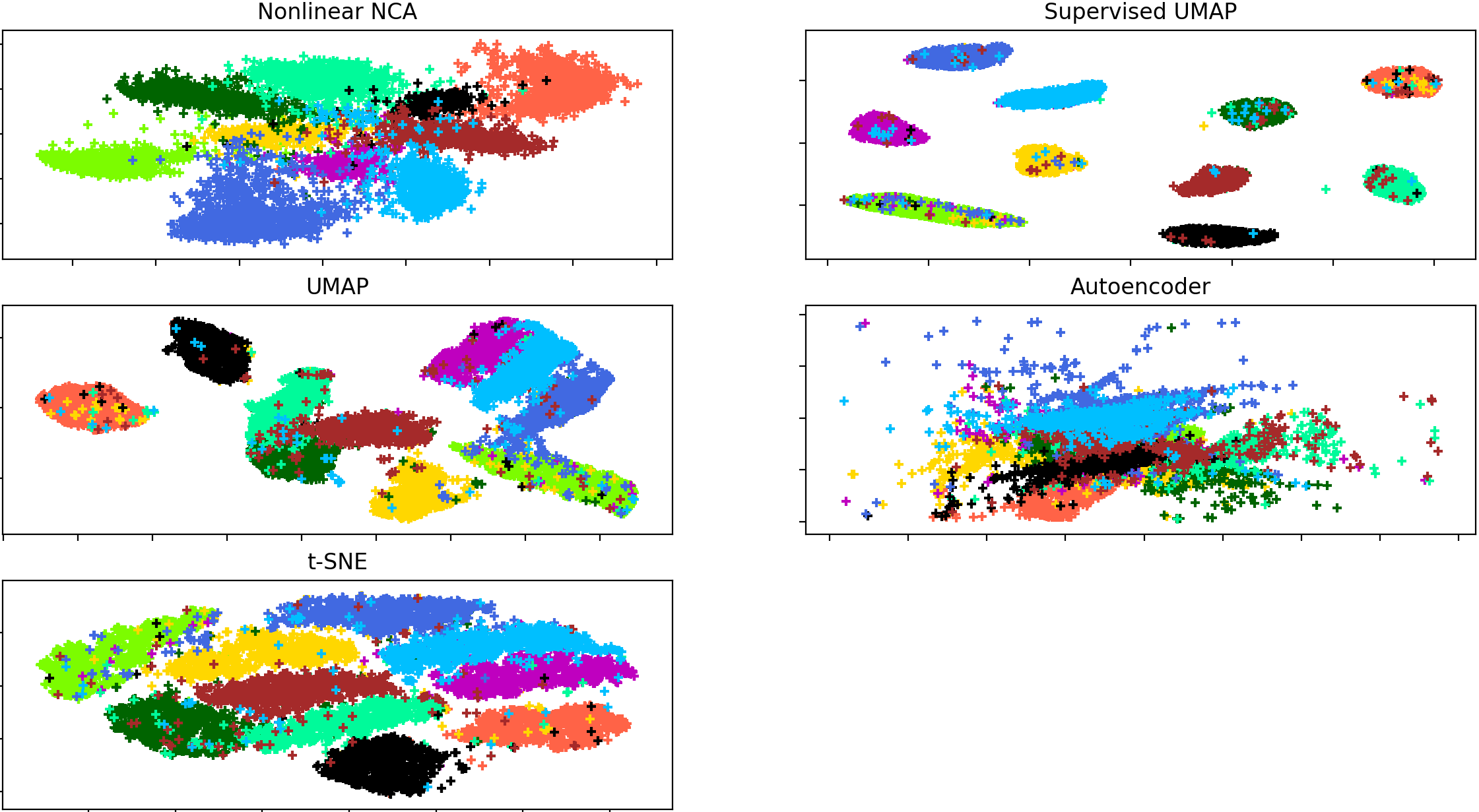}
	\caption{A comparison of the visualizations  of  10000 MNIST test digits by different  methods.}
	\label{fig:MNIST_2d_Tstdata_multiple_methods}
\end{figure}

\newpage
%Now we compare CE applied to the MNIST data with other supervised visualization methods.  
Figure \ref{fig:MNIST_2d_Tstdata_multiple_methods} shows the two-dimensional arrangement of 10,000 MNIST test set digits using nonlinear NCA, supervised UMAP, UMAP, autoencoder, and t-SNE.
It is apparent that CE is more similar to LE in terms of the relative positioning of the digits in 2D than these other supervised visualization  methods.  There are some similarities across all the models, e.g.,  0 and 6 are neighbors for all methods (except for SUMAP), as are the digits 4, 9, and 7. The separation among the different digit groups more prominent in CE than all the methods save for SUMAP, but as seen above, quantitatively, SUMAP has a higher error rate than CE. We note that unsupervised  UMAP, as well as t-SNE, both agree with CE that the digits 5 and 8 should be in the center.  However, in contrast, CE provides  a mapping that can be applied to streaming data.

\newpage
\subsubsection{USPS}

On USPS, centroid-encoder without pre-training has a prediction error slightly lower than HOPE but outperformed dt-MCML and dt-NCA by the margins of $1.09\%$ and $2.12\%$, respectively. 
Without pre-training, the centroid-encoder performed better than NNCA with pre-training and supervised-UMAP on both MNIST and USPS datasets.

On the USPS data, the top three models with pre-training are dt-MCML, centroid-encoder, and dG-MCML with the error rates of 2.46, 2.91, and 3.37, respectively. The error rate of centroid-encoder is better than NNCA and supervised-UMAP by a margin of $3.67\%$ and $3.26\%$, respectively. Like in MNIST, dt-NCA and dG-NCA performed relatively poorly compared to centroid-encoder. Again, autoencoder performed the worst among all the models.

\begin{figure}[ht!]
	\includegraphics[width=15.2cm, height=12.0cm]{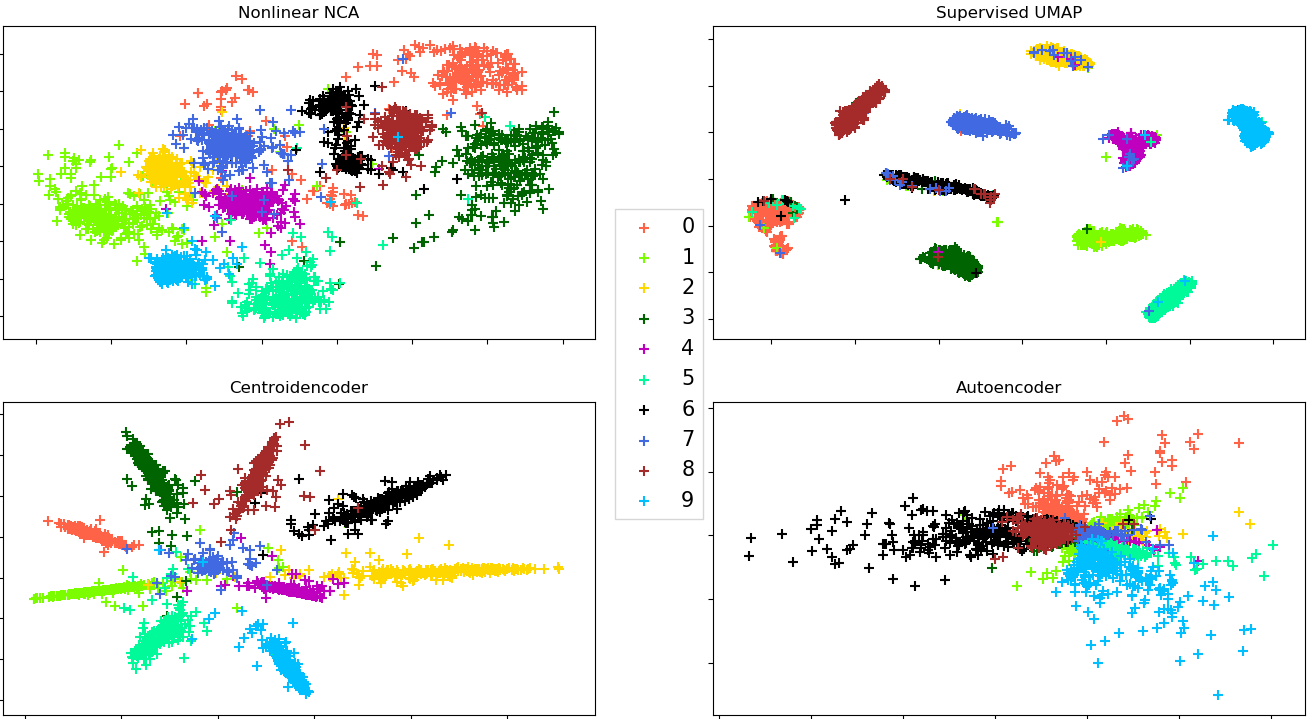}
	\caption{Two dimensional  plot of 3000 USPS test digits by different dimensionality reduction methods.}
	\label{fig:USPS_2d_Tstdata_multiple_methods}
\end{figure}

  In Figure \ref{fig:USPS_2d_Tstdata_multiple_methods}, we show the two-dimensional visualization of 3,000 test samples using nonlinear NCA, supervised UMAP, centroid-encoder, and autoencoder.    
 Like MNIST, the neighborhoods of CE and the other methods share many similarities.
Digits 4, 9, and 7 are still neighbors  in all the techniques but now 
sit more centrally.  Digits 8 and 6 are  now consistently neighbors across all methods.
  The within-class scatter of each digit is again the highest for autoencoder since it is unsupervised.
  Nonlinear NCA also has considerable variance across all the digit classes. Supervised UMAP has tighter clumping of the classes -- apparently
  an integral feature of SUMAP as well as UMAP. However, supervised UMAP has some misplaced digits compared to CE, which is evident from the error rate in Table \ref{table:MNIST_USPS_Result_W_pretr}.
  These observations are  consistent with MNIST visualizations.    SUMAP emphasizes the  importance of local structure over the global structure of the data. 
  In contrast to CE, Autoencoder doesn't provide any meaningful information about the neighborhood structure.

\newpage
\subsubsection{Letter, Landsat and Phoneme Data Sets}
Here we compare centroid-encoder with another suite of  supervised methods including, SNeRV, PE, S-Isomap, MUHSIC, MRE, and NCA. Table \ref{table:CEvsSNeRV} shows the results on three benchmarking data sets, including  UCI Letter, Landsat, and Phoneme dataset.  On Landsat data, we include UMAP and SUMAP for additional comparison. CE produces the smallest prediction error
on Landsat and Phoneme data sets  and is ranked second
on the Letter data set. On Landsat dataset, the top three models are CE, SUMAP and UMAP. 
On the Phoneme dataset, the centroid-encoder outperforms the second-best model, which is S-Isomap, by a margin of $1.9\%$. Notably, the variance of errors of centroid-encoder ($1.33$) is better than the S-Isomap ($5.74$). On the Letter dataset, the centroid-encoder achieves the error rate of $23.82\%$, which is the second-best model after SNeRV ($\lambda=0.1$), although the variance of the result of centroid-encoder is better than SNeRV by a margin of $1.47$. It's also noteworthy that the variance of errors for  centroid-encoder is the lowest {\it in every case}.

\begin{table}[ht!]	
	\centering
	\begin{tabular} {|c|c|c|c|}	% (3 columns)
		% start header
		\hline\hline 	% Makes 2 fancy lines
		 %\multicolumn{1}{|c|}{Method} & \multicolumn{3}{c|} {Dataset}\\
		 \multirow{2}{*}{Method} & \multicolumn{3}{c|} {Dataset}\\\cline{2-4}		
			                                       & Letter & Landsat & Phoneme \\		
		\hline
		CE & $23.82 \pm 3.13$ & $\textbf{8.97} \pm \textbf{2.55}$ & $\textbf{7.52} \pm \textbf{1.33}$ \\
		SUMAP & NA & $9.68 \pm 3.12$ & NA \\
		UMAP & NA & $11.89 \pm 2.83$ & NA \\
		SNeRV $\lambda=0.1$ & $\textbf{22.96} \pm \textbf{4.6}$\ &$14.34 \pm 7.38$ & $9.43 \pm 7.79$ \\
		SNeRV $\lambda=0.3$ & $24.59 \pm 4.6$ & $13.93 \pm 6.97$ & $9.02 \pm 7.38$ \\
		PE  & $31.15 \pm 4.92$ & $14.75 \pm 8.20$ & $9.84 \pm 6.15$ \\
		S-Isomap  & $31.97 \pm 7.38$ & $15.16 \pm 9.02$ & $8.61 \pm 5.74$ \\
		NCA  & $62.30 \pm 5.74$ & $15.57 \pm 7.38$ & $20.49 \pm 5.73$ \\
		MUHSIC  & $79.51 \pm 4.92$ & $15.37 \pm 4.10$ & $14.75 \pm 4.1$ \\		
		MRE  & $90.98 \pm 7.38$ & $53.28 \pm 34.2$ & $45.08 \pm 18.03$ \\
		\hline \hline
	\end{tabular}	
	\caption{Classification error (\%) of $k$-NN ($k$=5) on the 2D embedded data by different supervised embedding methods on Letter, Landsat and Phoneme data. Average misclassifications over ten-fold cross-validation are shown along with standard deviation. Results of methods other than centroid-encoder, UMAP and SUMAP are reported from \cite{Venna:2010:IRP:1756006.1756019}. NA means result is not available.}
	\label{table:CEvsSNeRV}
\end{table}

\begin{figure}[pt!]
	\vspace{-0.75cm}    
	\begin{subfigure}[b]{1.00\textwidth}
		\centering
		\includegraphics[width=12.0cm, height=9.65cm]{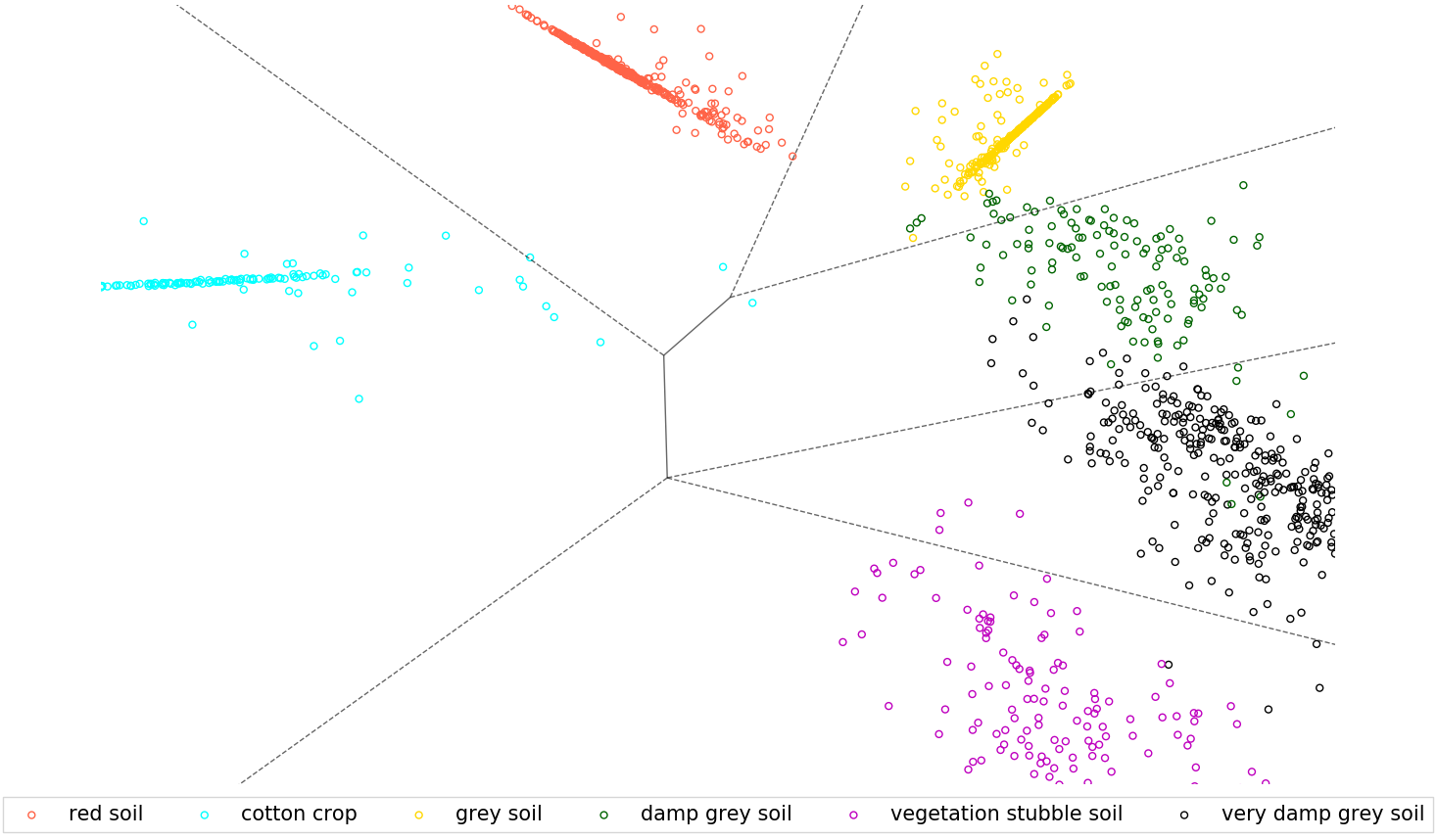}        
		\caption{Two-dimensional plot of Landsat training samples.}
		\vspace{0.5cm}
	\end{subfigure}    
	\begin{subfigure}[b]{1.00\textwidth}
		\centering
		\includegraphics[width=12.25cm, height=9.65cm]{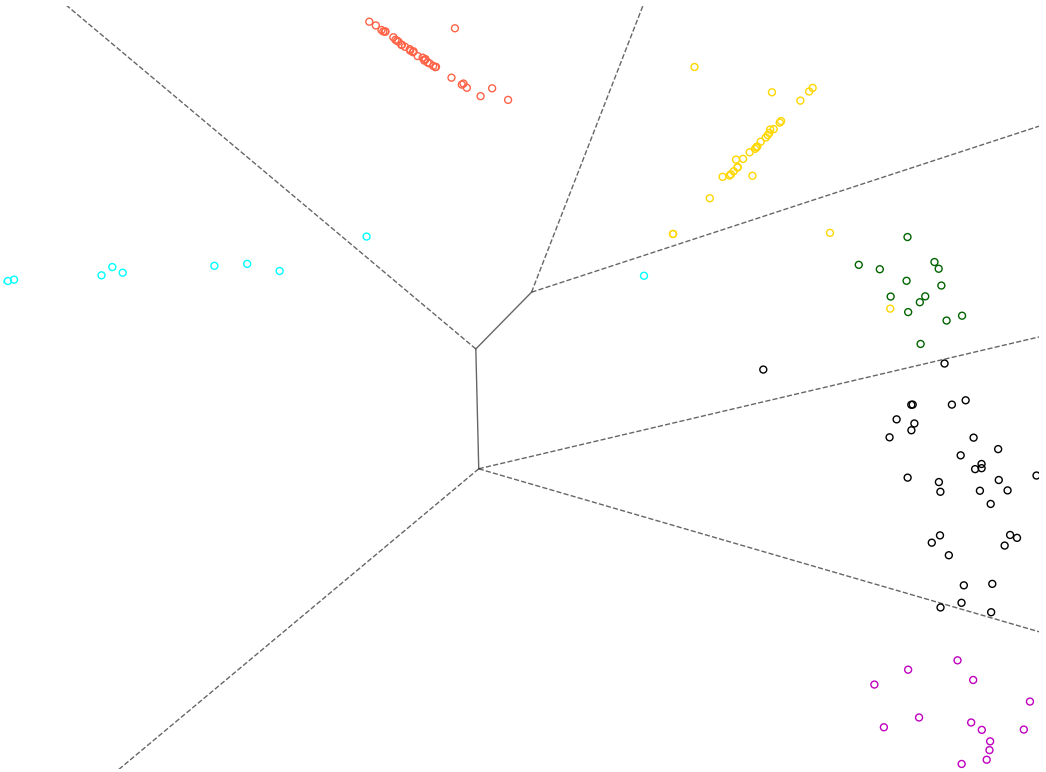}
		\caption{Two-dimensional plot of Landsat test samples.}  
	\end{subfigure}   
	\caption{Voronoi cells of low-dimensional Landsat data using a $16\rightarrow[250,150,2,150,250]\rightarrow16$ centroid-encoder. Voronoi regions for each soil type are formed from the training set.}
	\label{LandSatCE}
\end{figure}

\begin{figure}[pt!]	
	\begin{subfigure}[b]{1.00\textwidth}
		\centering
		\includegraphics[width=12.0cm, height=9.65cm]{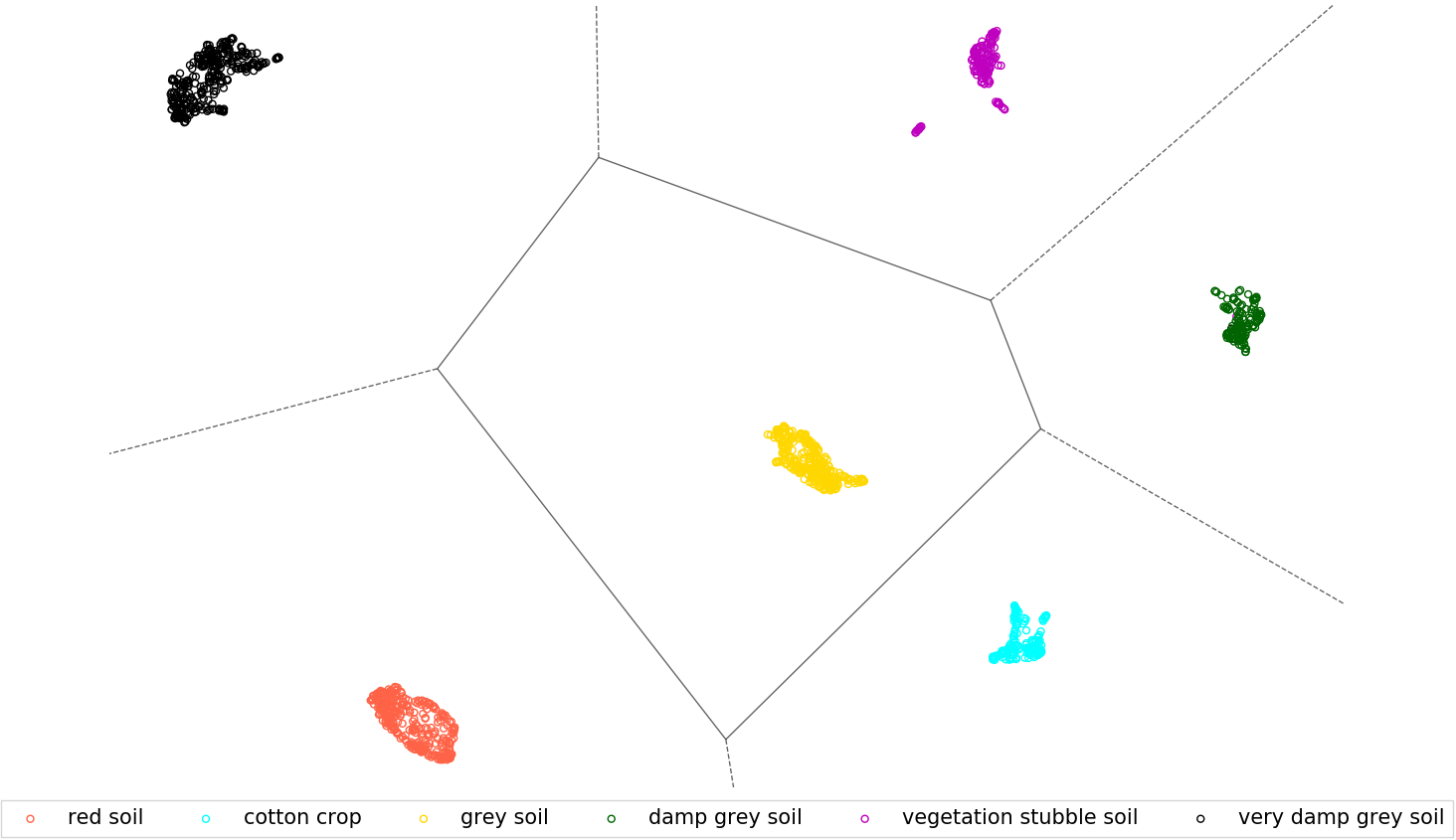}        
		\caption{Two-dimensional plot of Landsat training samples.}		
	\end{subfigure}    
	\begin{subfigure}[b]{1.00\textwidth}
		\centering
		\includegraphics[width=12.25cm, height=9.65cm]{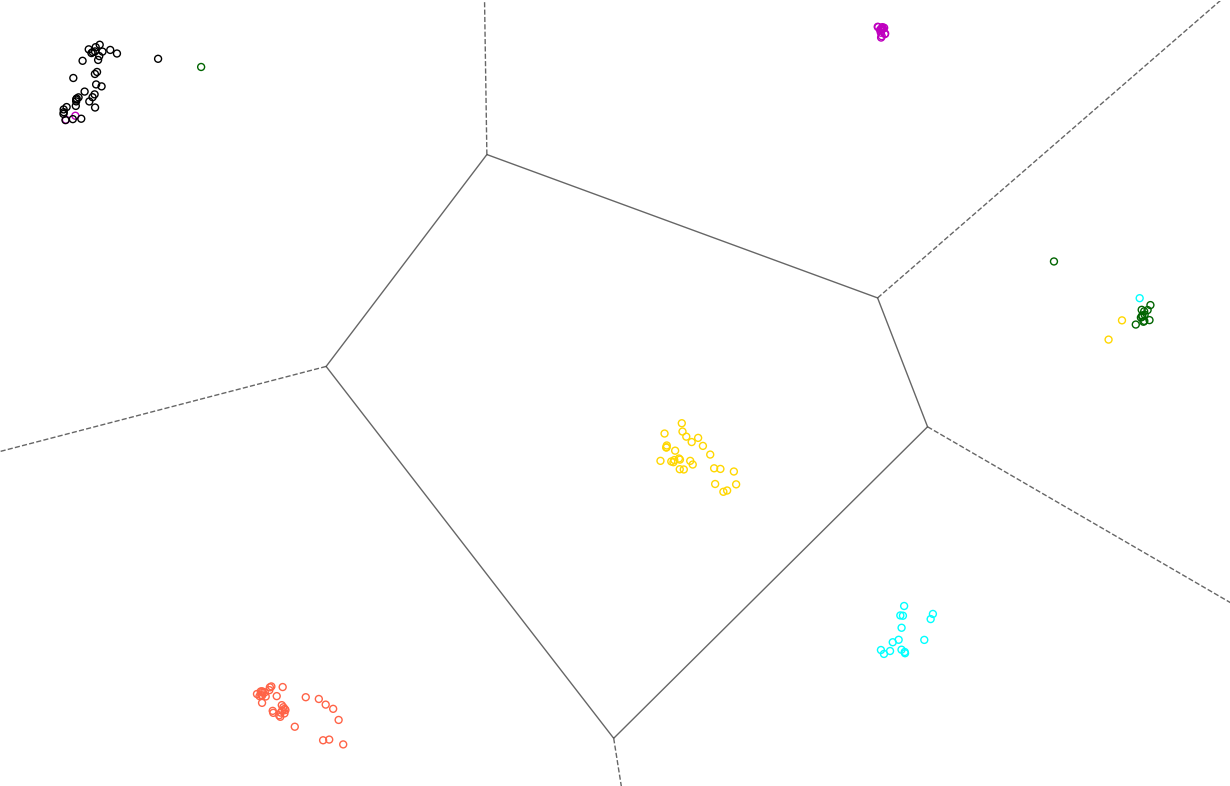}
		\caption{Two-dimensional plot of Landsat test samples.}  
	\end{subfigure}   
	\caption{Voronoi cells of low-dimensional Landsat data using Supervised UMAP. Voronoi regions for each soil type are formed from the training set.}
	\label{LandSatSUMAP}
\end{figure}

The visualization of the two-dimensional embedding using centroid-encoder on Landsat is shown in Figure
\ref{LandSatCE}.  We might conclude that the water content
in the samples is causing scatter in the 2D plots and decreases as we 
proceed counter-clockwise.  The test samples encode essentially the same 
structure. In Figure \ref{LandSatSUMAP}, we show the visualization using SUMAP. We see tighter clusters, but there is no difference between the damp and non-damp soil types.    The neighborhood relationships, demonstrated by Laplacian Eigemaps, also show this transition by dampness 
consistent with CE. Damp and very damp soils are, however, not neighbors using SUMAP, making the visualization less informative.

\newpage
\subsubsection{Iris, Sonar and USPS (revisited) Data Sets}
In our last experiment, we compare supervised PCA and kernel supervised
PCA to centroid-encoder. Table \ref{table:KSPCA_SPCA_CE_Error} contains the classification error using a 5-NN classifier, showing that centroid-encoder outperforms SPCA and KSPCA by considerable margins
 by this objective measure.

\begin{figure}[ht!]
\begin{subfigure}[t]{0.3\textwidth}   
    \includegraphics[width=5.0cm, height=5.0cm]{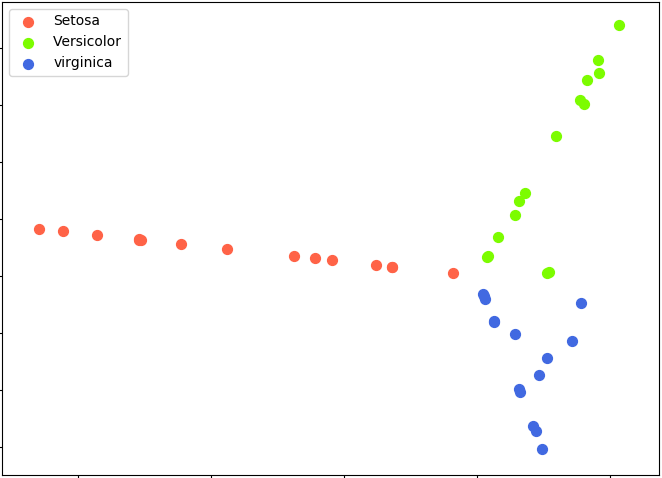}
\caption{KSPCA}
\label{fig:Iris_KSPCA}
\end{subfigure}\hfill
\begin{subfigure}[t]{0.3\textwidth}
  \includegraphics[width=5.0cm, height=5.0cm]{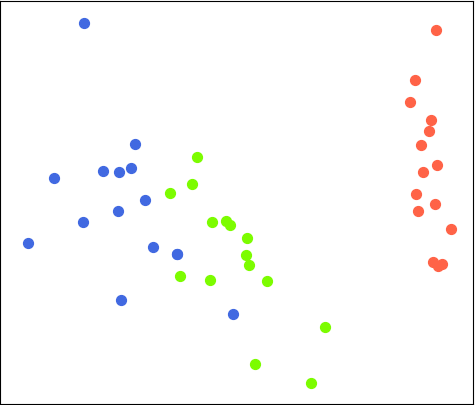}
\caption{SPCA}
\label{fig:Iris_SPCA}
\end{subfigure}\hfill
\begin{subfigure}[t]{0.3\textwidth}
    \includegraphics[width=5.0cm, height=5.0cm]{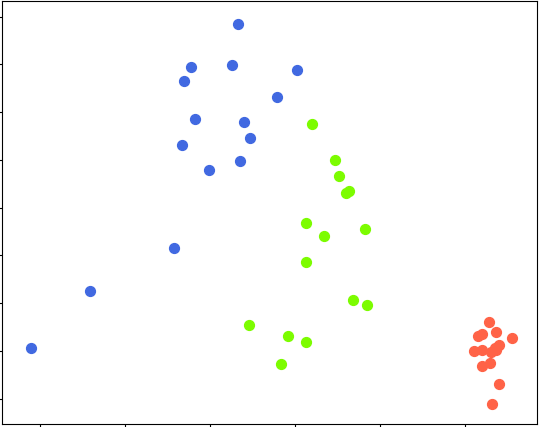}
\caption{Centroid-encoder}
\label{fig:Iris_CE}
\end{subfigure}

\begin{subfigure}[t]{0.3\textwidth}
    \includegraphics[width=5.0cm,height=5.0cm]{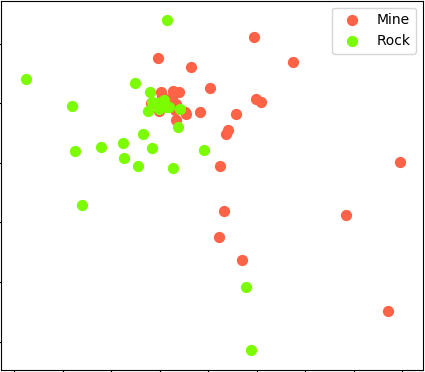}
\caption{KSPCA}
\label{fig:Sonar_KSPCA}
\end{subfigure}\hfill
\begin{subfigure}[t]{0.3\textwidth}
  \includegraphics[width=5.0cm,height=5.0cm]{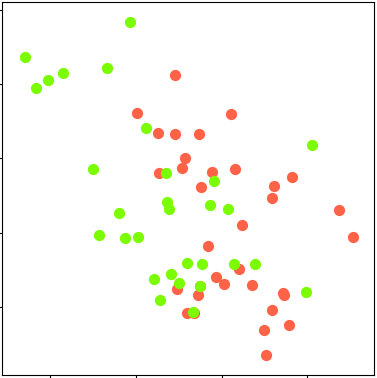}
\caption{SPCA}
\label{fig:Sonar_SPCA}
\end{subfigure}\hfill
\begin{subfigure}[t]{0.3\textwidth}
    \includegraphics[width=5.0cm,height=5.0cm]{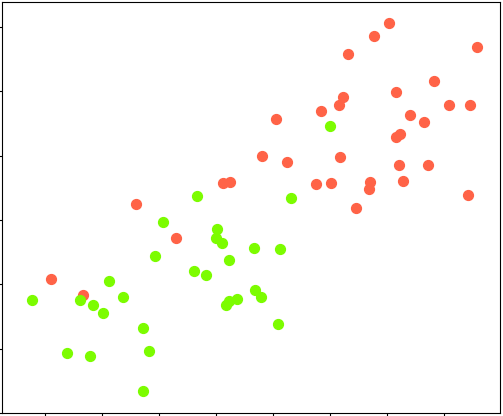}
\caption{Centroid-encoder}
\label{fig:Sonar_CE}
\end{subfigure}

\begin{subfigure}[t]{0.3\textwidth}
    \includegraphics[width=5.0cm,height=5.0cm]{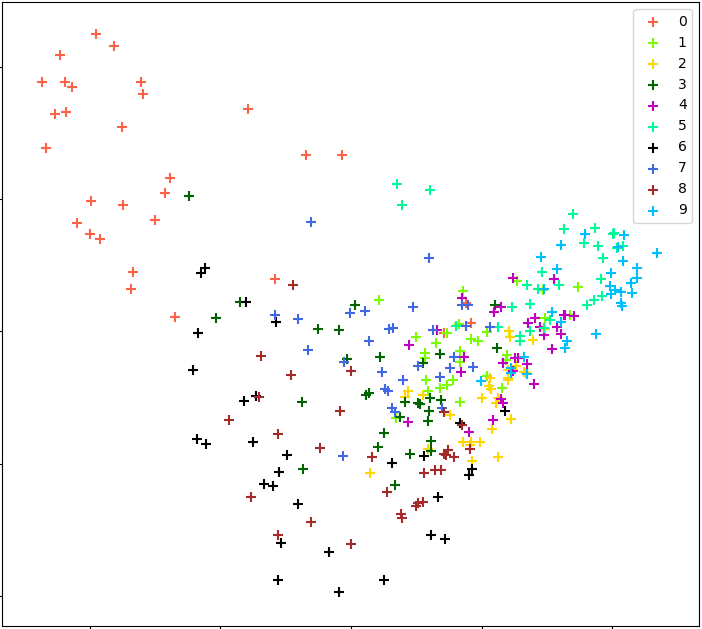}
\caption{KSPCA}
\label{fig:USPS_1000_KSPCA}
\end{subfigure}\hfill
\begin{subfigure}[t]{0.3\textwidth}
  \includegraphics[width=5.0cm,height=5.0cm]{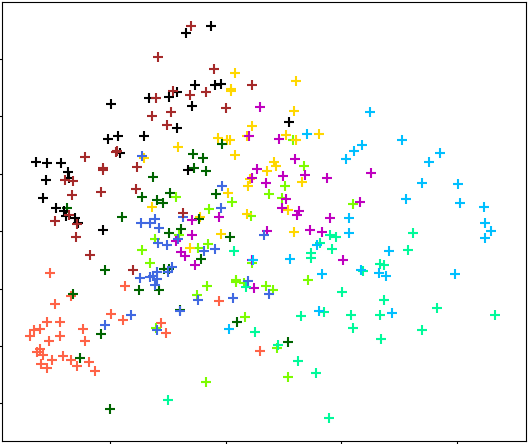}
\caption{SPCA}
\label{fig:USPS_1000_SPCA}
\end{subfigure}\hfill
\begin{subfigure}[t]{0.3\textwidth}
    \includegraphics[width=5.0cm,height=5.0cm]{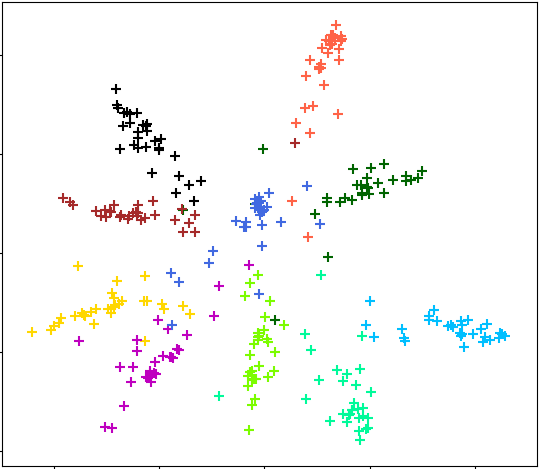}
\caption{Centroid-encoder}
\label{fig:USPS_1000_CE}
\end{subfigure}
\caption{Visualization of Iris (top row), Sonar (middle row) and USPS (bottom row) data using different dimensionality reduction techniques.}
\label{fig:Iris_Sonar_USPS_2d_Tr_Tstdata_multiple_methods}
\end{figure}

In the top row of Figure \ref{fig:Iris_Sonar_USPS_2d_Tr_Tstdata_multiple_methods}, we present the two-dimensional visualization of Iris test data. Among the three methods, centroid-encoder and KSPCA produce better separation compared to SPCA. SPCA can't separate the Iris plants Versicolor and Virginica in 2D space. The dispersion of the three Iris categories is relatively higher in KSPCA and SPCA compared to centroid-encoder. Surprisingly in KSPCA, three classes are mapped on three separate lines.

The two-dimensional visualization of Sonar test samples is shown in the middle row of Figure \ref{fig:Iris_Sonar_USPS_2d_Tr_Tstdata_multiple_methods}. None of the methods can completely separate the two classes. The embedding of KSPCA is slightly better than SPCA, which doesn't separate the data at all. In centroid-encoder, the two classes are grouped relatively far away from each other. Most of the rock samples are mapped at the bottom-left of the plot, whereas the mine samples are projected at the top-right corner.

The visualization of USPS test data is shown at the bottom of Figure \ref{fig:Iris_Sonar_USPS_2d_Tr_Tstdata_multiple_methods}. Both SPCA and KSPCA are unable to separate the ten classes, although KSPCA separates digit 0s from the rest of the samples. In contrast, the centroid-encoder puts the ten digit-classes in 2-D space without much overlap. Qualitatively, the embedding of centroid-encoder is better than the other two.

\begin{table}[ht!]	
	\centering
	\begin{tabular} {|c|c|c|c|}	% (3 columns)
		% start header
		\hline\hline 	% Makes 2 fancy lines
		 \multirow{2}{*}{Method} & \multicolumn{3}{c|} {Dataset}\\
		 \cline{2-4}		
			                                       & \multicolumn{1}{c|} {IRIS} &  \multicolumn{1}{c|} {Sonar}&  \multicolumn{1}{c|} {USPS} \\

		\hline
		Centroidencoder  & $\textbf{3.29} \pm \textbf{2.10}$ & $\textbf{14.24} \pm \textbf{2.99}$& $\textbf{12.16} \pm \textbf{2.01}$ \\
		KSPCA & $5.16\pm 4.56$ & $21.06\pm 5.38$ & $51.63 \pm 2.11$ \\
		SPCA & $66.67\pm 0.00$ & $49.06\pm 5.04$ & $64.95 \pm 1.94$\\
		\hline \hline
	\end{tabular}	
	\caption{$k$-NN ($k=5$) error (\%) on the 2D embedded data by different supervised embedding methods on Iris, Sonar and USPS data. Note that given the limitations of
	KSPCA we have restricted the USPS data set to have
1000 total samples.}
	\label{table:KSPCA_SPCA_CE_Error}
\end{table}

\newpage
\section{Discussion and Analysis of Results}
\label{discussion}
The experimental results in Section \ref{quantviz} establish that centroid-encoder performs competitively, i.e., it is the best, or near the best, on a diverse set of data sets.  {The algorithms that produce similar 
classification rates require the computation of distance matrices and are hence more expensive than centroid-encoders.} Here we analyze the CE algorithm in more detail to discover why this improved performance is perhaps not surprising. This has to do, at least in part, with the fact
that centroid-encoder,  implementing a nonlinear mapping, is actually able to capture more
variance than PCA, and this is particularly important in lower dimensions.

\subsection{Variance Plot to Explain the Complexity of Data}
\begin{figure}[!ht]
        \centering        
        \includegraphics[width=15.0cm,height=7.65cm]{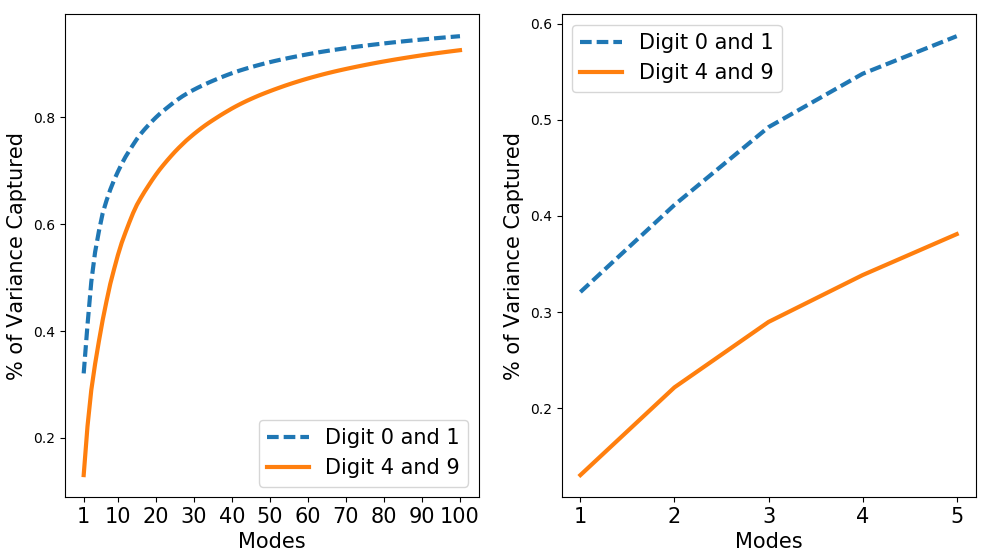}        
        \caption{Variance plot using two different subsets (subset1: all the digits 0 and 1, subset2: all the digits 4 and 9) of MNIST data. Left: comparison of variance plot of the two subsets using the first 100 dimensions. Right: a blowup of the plot on the left showing the  \% of variance captured in low dimensions.  }
        \label{fig:EnergyPlotMNIST}
\end{figure}

\begin{figure}
    \centering
    \begin{subfigure}[b]{0.45\textwidth}
        \includegraphics[width=7.25cm, height=6.25cm]{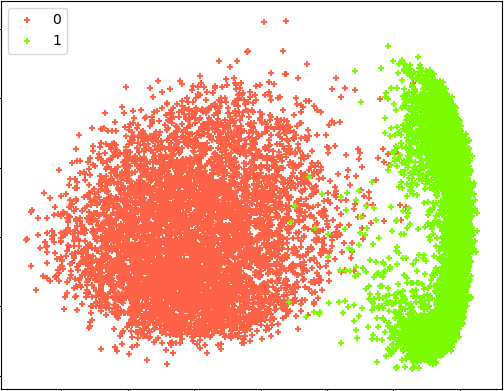}
        \caption{Subset 1: digit 0 and 1}
        \label{fig:Digit0_1}
    \end{subfigure}
    ~ %add desired spacing between images, e. g. ~, \quad, \qquad, \hfill etc. 
      %(or a blank line to force the subfigure onto a new line)
    \begin{subfigure}[b]{0.45\textwidth}
        \includegraphics[width=7.25cm, height=6.25cm]{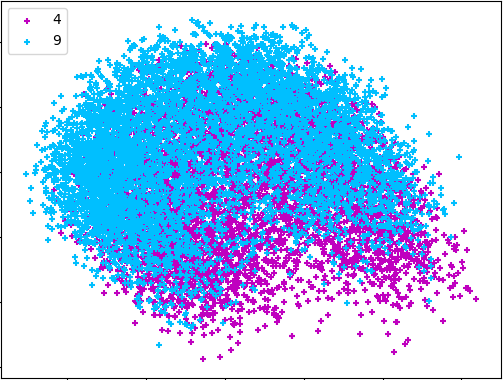}
        \caption{Subset 2: digit 4 and 9}
        \label{fig:Digit4_9}
    \end{subfigure}
    \caption{Two dimensional embedding using PCA for subset 1 and subset 2. }\label{fig:2DPCAofDigit0_1andDigit4_9}
\end{figure}

PCA captures the maximum variance over the class of orthogonal transformations \cite{kirby_wiley2}.
The total variance captured as a function of the number
of dimensions is a measure of the complexity of a data set.
We can use variance to establish, e.g., that the digits 0 and 1 are less
complex than the digits 4 and 9.  To this end, we use two subsets from the MNIST training set: the first one contains all the digits 0 and 1, and the second one has all the digits 4 and 9.  Next, we compute the total variance captured as a function of dimension; see Figure \ref{fig:EnergyPlotMNIST}. 
Clearly, the variance is more spread out across the dimensions
for the digits 4/9. In Figure \ref{fig:2DPCAofDigit0_1andDigit4_9}, we show the visualization of these two subsets using PCA. We see that the 0/1 samples in subset1 are well separated in two-dimensional space, whereas the data in subset2 are clumped together. This lies in the fact that in low dimension (2D) PCA captures more variance from subset1 (around 41\%) as compared to subset2 (about 22\%).  We can say that subset2 is more complex than subset1.

\subsection{Centroid-encoder and Variance}
\begin{figure}[!ht]
        \centering        
        \includegraphics[width=15.0cm,height=7.15cm]{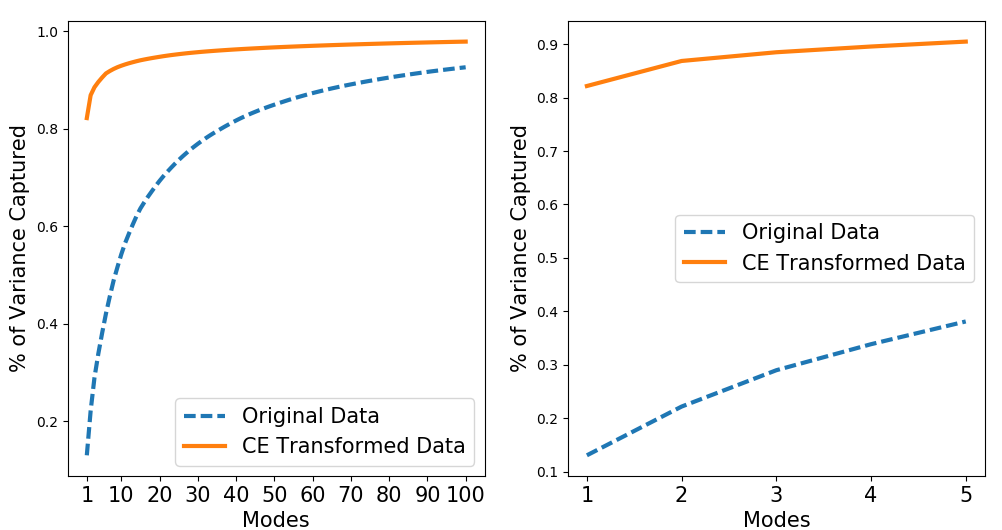}
        \caption{Variance plots for the 4/9 digit data. Left: comparison of variance plot of  the original data and CE transformed data using the first 100 dimensions.  Right: a blowup of the plot on the left showing the  \% of variance captured in low dimensions.}
        \label{fig:EnergyPlotCETransformedMNIST}
\end{figure}

\begin{figure}[!ht]
    \centering
    \begin{subfigure}[b]{0.45\textwidth}        
        \includegraphics[width=7.25cm, height=6.25cm]{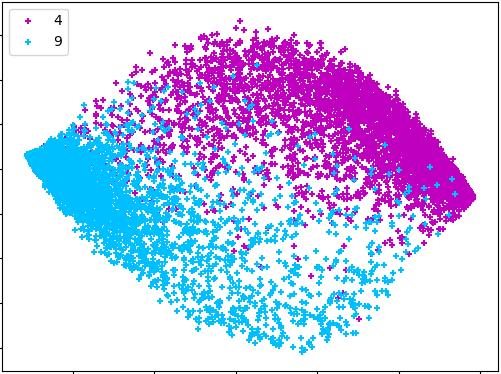}
        \caption{Digits 4 and 9 from training set.}
        \label{fig:TrDigits4_9_CE}
    \end{subfigure}
    ~ %add desired spacing between images, e. g. ~, \quad, \qquad, \hfill etc. 
      %(or a blank line to force the subfigure onto a new line)
    \begin{subfigure}[b]{0.45\textwidth}
        \includegraphics[width=7.25cm, height=6.25cm]{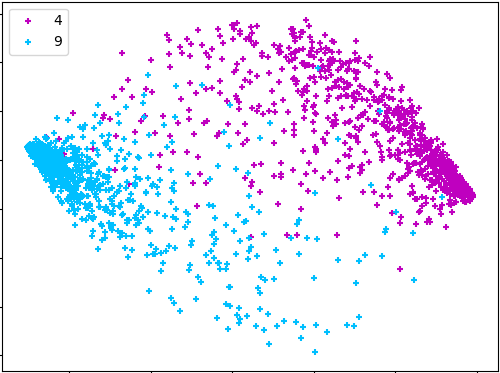}
        \caption{Digits 4 and 9 from test set.}
        \label{fig:TstDigits4_9_CE}
    \end{subfigure}
    \caption{Two dimensional embedding using PCA on the CE-transformed data. }
    \label{fig:2DPCAofCE-transformedDigit4_9}
\end{figure}

In this section, we demonstrate how centroid-encoder
maps a higher fraction of the variance of the data to lower dimensions. 
Again,  consider the subset of digits 4 and 9 from the experiment 
above. We use all the digits 4 and 9 from the MNIST test set as a validation set. We trained a centroid-encoder with the architecture $784\rightarrow[784]\rightarrow784$ where we used hyperbolic tangent (`tanh') as the activation function. We keep the number of nodes the same in all the three layers. After the training, we pass the training and test samples through the network and capture the hidden activation which we call CE-transformed data. We show the variance plot and the two-dimensional embedding by PCA on the CE-transformed data. Figure \ref{fig:EnergyPlotCETransformedMNIST} compares the variance plot between the original data and the CE-transformed data. In the original data, only 22\% of the total variance is captured, whereas about 89\% of the total variance is captured in CE-transformed data. It appears that the nonlinear transformation of centroid-encoder puts the original high dimensional data in a relatively low dimensional space. Visualization using PCA of CE-transformed data (both training and test) is shown in Figure \ref{fig:2DPCAofCE-transformedDigit4_9}. This time, PCA separates the classes in two-dimensional space. Comparing Figure \ref{fig:2DPCAofCE-transformedDigit4_9} with Figure \ref{fig:Digit4_9}, it appears that the transformation by centroid-encoder is helpful for low dimensional visualizations.

\subsection{Computational Complexity and Scalability}
The computational complexity of the centroid-encoder is effectively the same as the autoencoder.   The advantage of centroid-encoder over
many other data visualization techniques is that it is not necessary to 
compute pairwise distances.  If everything else is fixed, we can view
centroid-encoder as scaling  linearly
with the number of data points.   Techniques that require distance
matrices scale quadratically with the number of data points and can be 
prohibitively slow as the data set size increases.   
Methods based on distance matrices include
 NNCA, dt-NCA, dG-NCA, dt-MCML, and dG-MCML,
as well as the spectral methods including  MDS, UMAP, Laplacian Eigenmaps, Isomap and their supervised counterparts. The only overhead of centroid-encoder is the calculation of centroids of each class, but this is also linear with the number of classes.

\section{Conclusion}
\label{conc}
In this paper, we presented the centroid-encoder for the task of
data visualization when class labels are available.  We compared CE to state-of-the-art
techniques and showed advantages empirically over other methods.  Our examples illustrate that CE captures the topological structure, i.e.,  neighborhoods of the data,
comparable to methods, but   with lower cost and generally better prediction error.
 These experiments include comparisons to the unsupervised Laplacian eigenmaps, UMAP, pt-SNE, autoencoders;  the supervised UMAP, SPCA, KSPCA 
as well as  the MCML and NCA class of algorithms.  The algorithms that have the most similar prediction errors when compared to CE  require the computation of distances between all pairs of points.
We demonstrated that the 2D embedding of centroid-encoder produces competitive, if not optimal,  prediction errors. We also showed empirically that the model globally captures  high statistical variance relative to  optimal linear transformations, i.e., more than PCA.

In addition to capturing the global topological structure of the data in low-dimensions,  CE exploits data labels to minimize the within-class variance. This improves the localization of the mapping such that points are mapped more faithfully to their Voronoi regions in low-dimensions. CE also provides a mapping model  that can be applied to new data without any retraining being necessary. This is in contrast to the spectral methods  described here that require the entire training and testing set for computing the embedding eigenvectors.

In the future, we plan to extend the model in unsupervised and semi-supervised learning. 

\section{Acknowledgement}
The authors thank Shannon Stiverson and Eric Kehoe to review the article thoroughly and giving us their feedback to make the material better. We also thank Manuchehr Aminian to make sure that the code in GitHub is working correctly.

\newpage
\bibliographystyle{unsrt}
\bibliography{VisualizationUsingCE,complete6,KirbyNov2019,books}
%\bibliography{references}  %%% Remove comment to use the external .bib file (using bibtex).
%%% and comment out the ``thebibliography'' section.
\end{document}